\documentclass[journal]{IEEEtran}
\usepackage{amsmath,amsfonts}
\usepackage{algorithmic}
\usepackage{algorithm}
\usepackage{array}
\usepackage[caption=false,font=footnotesize,labelfont=sf,textfont=sf]{subfig}
\usepackage{textcomp}
\usepackage{stfloats}
\usepackage{verbatim}
\usepackage{graphicx}
\usepackage{cite}
\usepackage{microtype}
\usepackage{xcolor}
\newcommand{\gravabrand}{%
  \raisebox{-0.28\height}{\shortstack[c]{%
  \includegraphics[height=7mm]{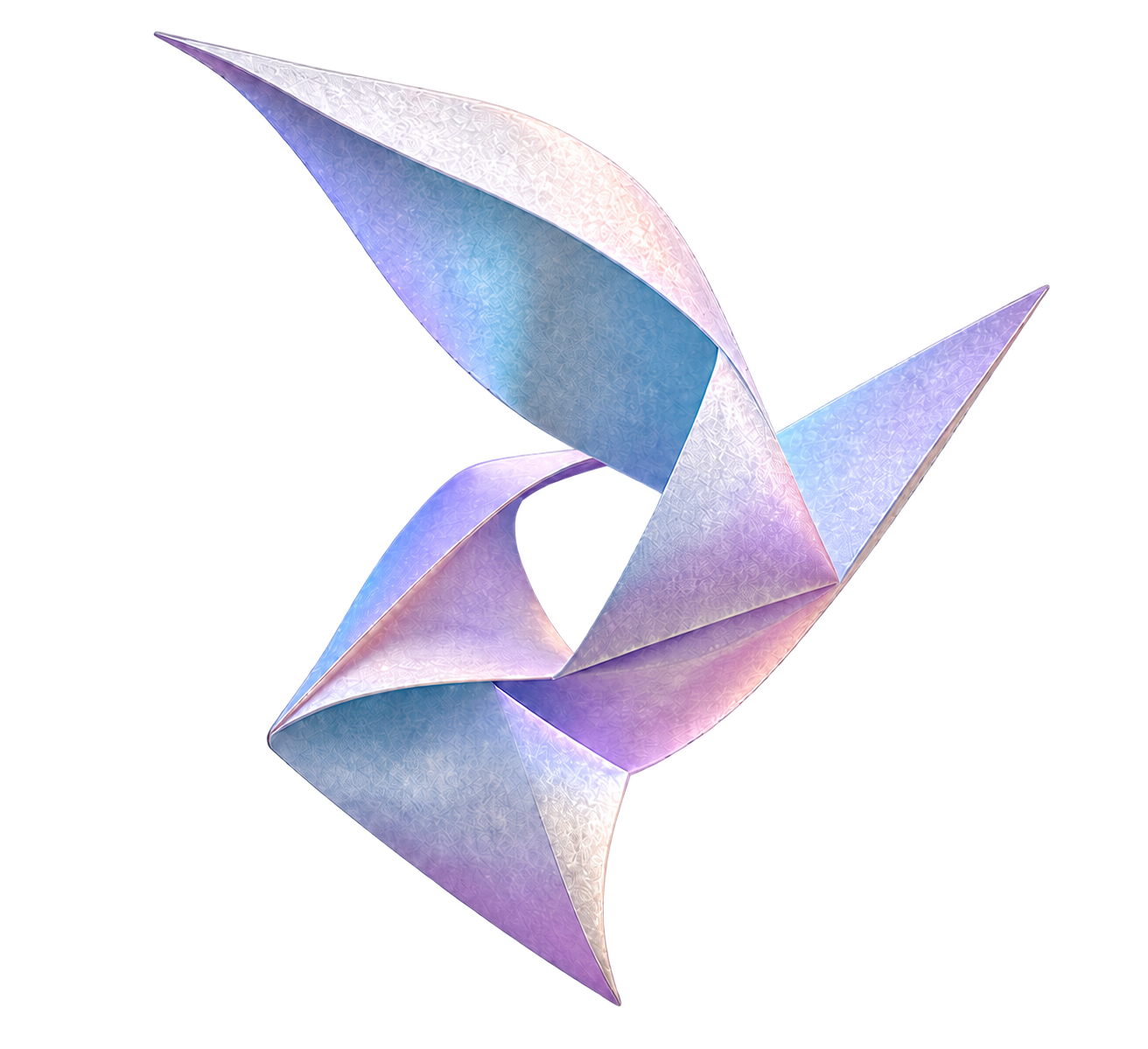}\\[-0.5mm]
  \scalebox{0.45}{{\sffamily\mdseries\fontsize{10}{11}\selectfont
   \textcolor[HTML]{34495E}{\textls[200]{GRAVA}}}}}}%
}

\usepackage{tabularx}
\usepackage{booktabs}
\usepackage{adjustbox}
\usepackage{listings}
\usepackage{placeins}
\usepackage[most]{tcolorbox}
\usepackage[T1,OT1]{fontenc}
\usepackage{hyperref} 
\usepackage{orcidlink}

\begin{document}

\bstctlcite{grava:bstcontrol}

\hypersetup{pdftitle={GRAVA: Grounded Reasoning-to-Action Representation and Learning for Autonomous Driving}}
\title{\mbox{\gravabrand\hspace{0.35em}GRAVA:} Grounded Reasoning-to-Action Representation and Learning for \\ Autonomous Driving}

\author{Xiao Liu~\orcidlink{0009-0005-8242-8127}, Haoyu Li~\orcidlink{0009-0001-5494-0220}, Jianghao Leng~\orcidlink{0000-0002-3577-7990}, Lin Wang~\orcidlink{0000-0002-7485-4493},~\IEEEmembership{Member,~IEEE,} Chao Sun~\orcidlink{0000-0002-9324-0892},~\IEEEmembership{Member,~IEEE}

\thanks{Xiao Liu, Jianghao Leng, and Chao Sun are with the National Engineering Research Center of Electric Vehicles, Beijing Institute of Technology, Beijing 100081, China; the Shenzhen Automotive Research Institute, Beijing Institute of Technology, Shenzhen 518118, China; and Shenzhen Jiguangzhijie Technology Co., Ltd., Shenzhen 518118, China (e-mail: bitliuxiao@gmail.com; lengjianghao@szari.ac.cn; chaosun@bit.edu.cn). Corresponding author: Chao Sun.}
\thanks{Haoyu Li is with the National Engineering Research Center of Electric Vehicles, Beijing Institute of Technology, Beijing 100081, China, and the School of Electrical and Electronic Engineering, Nanyang Technological University, Singapore (e-mail: rabbit.yujixyz@gmail.com).}
\thanks{Lin Wang is with the School of Electrical and Electronic Engineering, Nanyang Technological University, Singapore 639798 (e-mail: linwang@ntu.edu.sg).}
\thanks{Code repository: \url{https://github.com/AhernResearch/grava}.}
\thanks{This work was supported by Shenzhen Science and Technology Program (ZDCY20250901103002003).}

        }

\markboth{SUBMITTED TO IEEE TRANSACTIONS ON PATTERN ANALYSIS AND MACHINE INTELLIGENCE, 2026}%
{Liu \MakeLowercase{\textit{et al.}}: GRAVA: Grounded Reasoning-to-Action Representation and Learning for Autonomous Driving}

\maketitle

\begin{abstract}
Driving vision-language-action (VLA) models increasingly incorporate reasoning before action prediction, yet their intermediate reasoning is often weakly grounded in physical scene evidence and inconsistently connected to executable behavior. We present GRAVA, a framework centered on Grounded Reasoning-to-Action (GRA) that integrates grounding, reasoning, and action generation within a single autoregressive stream. GRA binds action-relevant linguistic references to 2D visual regions and ego-centric physical states, while organizing object-specific interactions and driving decisions in a trajectory-anchored typed graph. Its action-relevant structure is serialized into grounded reasoning, with bounding-box tokens and physical quantities providing evidence for interaction and decision statements. A single VLM generates this reasoning followed immediately by a compact Executable Planner action, which is deterministically decoded into a continuous trajectory. An agentic GRA data construction pipeline combines forward scene grounding with backward trajectory anchoring to derive consistent cognition and planning supervision from the same graph. Using this pipeline, we construct GR-NavSim, extending the nuPlan dataset with 2.2M grounded question-answer pairs and 70K GRA reasoning traces. Centered on the GRA representation, we propose a progressive training strategy that develops grounded cognition through pre-training, establishes the reasoning-to-action interface through lightweight imitation, and further improves driving behavior through reinforcement learning and exploration. Using only about 60\% of the available human driving demonstrations for action supervision, GRAVA-8B achieves state-of-the-art performance among purely autoregressive driving models on the full NAVSIM benchmark. On an internal long-tail benchmark, full GRA improves key-object compliance and Closed-loop Driving Score by 19.3\% and 20.5\% relative to action-only prediction, respectively. These results demonstrate the benefit of preserving action-relevant physical evidence from grounded reasoning through executable action generation.
\end{abstract}



\section{Introduction}
\label{sec:introduction}

\begin{figure*}[t]
  \centering
  \includegraphics[width=\textwidth]{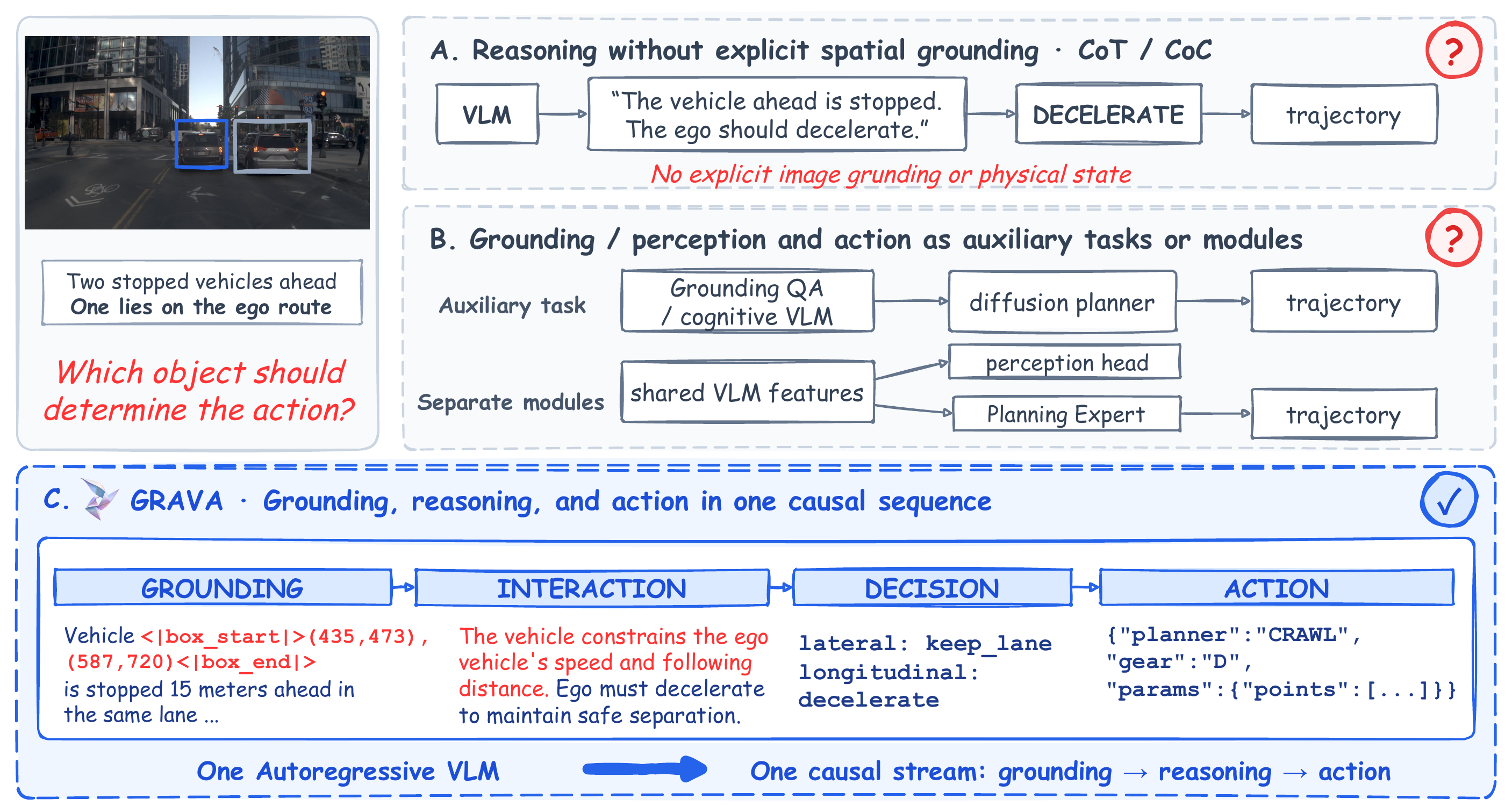}
  \caption{Grounding and reasoning-to-action gaps in driving VLAs. (A) Reasoning without explicit visual grounding. (B) Grounding separated from action generation through auxiliary tasks or modules. (C) GRAVA connects grounded evidence, interaction reasoning, decisions, and executable actions within one autoregressive stream.}
  \label{fig:gaps}
\end{figure*}

\IEEEPARstart{V}{ision-language-action (VLA)} models are emerging as a promising paradigm for autonomous driving by integrating multimodal scene understanding, reasoning, and action generation within a general-purpose generative model. Recent driving VLAs further introduce explicit intermediate reasoning before action prediction to support decision making in complex and long-tail scenarios~\cite{hwang2025emma,zhou2026opendrivevla,zhou2025autovla}. For driving, however, reasoning must identify not only the relevant traffic participants but also the physical states and interactions through which they constrain the ego behavior. A vehicle ahead, for example, affects the driving decision according to its lane occupancy, distance, and motion state. These scene-grounded relations must therefore remain explicitly connected to the action generated from the reasoning.

Despite this progress, existing driving VLAs still face two fundamental limitations, as illustrated in Fig.~\ref{fig:gaps}. First, we identify a \textbf{grounding gap (G1)} between linguistic reasoning and physical scene evidence. Action-relevant entities are often described through free-form language or high-level semantics without explicitly resolving them to visual regions and corresponding metric states. Consequently, a semantically plausible rationale may remain detached from the physical objects, distances, and motions that constrain the ego behavior. Second, we identify a \textbf{reasoning-to-action fragmentation gap (G2)}. Even when grounding is available, it is often introduced through dedicated perception or question-answering tasks, while reasoning and planning are learned from separate supervision or implemented through separate functional modules. The grounded objects and states established in one task are therefore not necessarily preserved as the evidence supporting subsequent interaction reasoning, decision making, and action generation. Addressing these limitations requires both explicit grounding of the reasoning content and continuity of that content through to executable behavior.

Establishing this continuity requires a representation that organizes scene evidence according to its role in the driving decision. Scene relevance is action dependent: visually salient objects are not necessarily those that constrain the demonstrated ego behavior, and different objects may impose distinct interaction requirements. Explicit visual references and physical states must therefore be connected to the object-specific interactions and decisions they support, rather than retained only as isolated perception attributes. This motivates an action-grounded representation that preserves the dependencies from physical scene evidence through reasoning to the final ego behavior.

To address this problem, we present \textbf{GRAVA}, a framework centered on the \textbf{Grounded Reasoning-to-Action (GRA)} representation. GRA organizes action-relevant scene context, grounded objects, physical states, interactions, decisions, and planning anchors within a trajectory-anchored typed graph. Each action-relevant linguistic referent is explicitly associated with its 2D visual region and ego-centric physical state, while object-specific interaction paths trace how this evidence contributes to the driving decision. The action-relevant graph structure is serialized into grounded reasoning in which bounding-box tokens and physical quantities remain associated with the interaction and decision statements they support. A single autoregressive VLM generates this reasoning followed immediately by a compact \textbf{Executable Planner} action, which is deterministically decoded into a continuous trajectory. GRA thus organizes the dependencies within the reasoning, while serialization and action generation carry them through the same autoregressive stream. Grounding becomes part of the reasoning used to generate the action, rather than an auxiliary task separated from downstream behavior.

Learning this grounded reasoning-to-action process requires supervision that preserves the same physical evidence and reasoning dependencies. GRAVA therefore employs a tool-augmented multi-agent annotation pipeline that combines forward scene grounding with backward trajectory anchoring. Forward grounding identifies candidate scene elements and establishes their visual and physical states, whereas backward anchoring starts from the expert trajectory and traces the demonstrated behavior back to the critical objects, interactions, and decisions that support it. Cognition-oriented question answering and planning-oriented reasoning-to-action supervision are derived as different projections of the resulting GRA graph, maintaining consistent object identities, physical states, and interaction semantics across the two forms of supervision. Grounded concepts learned through individual questions thus reappear within the reasoning that supports executable action generation.

Based on this supervision, GRAVA adopts a GRA-aligned learning process that progressively connects grounded cognition to executable behavior. GRA pre-training establishes reference resolution, physical-state understanding, interaction reasoning, and decision semantics. Lightweight planner warm-up then learns to generate executable actions as continuations of grounded reasoning, and verified self-distillation further stabilizes valid model-generated reasoning. Finally, an \textbf{Active RL Loop} optimizes complete reasoning-to-action sequences using trajectory-level driving rewards. By repeatedly selecting recoverable scenarios in which high-reward behavior is already reachable but not yet reliable, Active RL concentrates policy refinement on unresolved scenarios while retaining grounded reasoning and action as a single optimization unit.

Our contributions are threefold:
\begin{itemize}
\item We propose \textbf{Grounded Reasoning-to-Action (GRA)}, an action-grounded representation that binds linguistic references to 2D visual evidence and ego-centric physical states, while organizing their interactions and decisions in a trajectory-anchored typed graph. Its action-relevant structure is serialized into grounded reasoning followed by an Executable Planner action, preserving the progression from physical evidence to executable behavior within one autoregressive stream.

\item We develop a \textbf{GRA-aligned pre-training and post-training strategy} that progressively learns and optimizes this reasoning-to-action process. Grounded pre-training, planner warm-up, and verified self-distillation establish the transition from scene understanding to executable action generation, while the Active RL Loop refines complete sequences using driving outcomes on policy-dependent recoverable scenarios.

\item We develop an \textbf{agentic GRA data construction pipeline} and use it to build \textbf{GR-NavSim}, a vision-language alignment dataset based on nuPlan, comprising 2.2M grounded question-answer pairs and 70K GRA reasoning traces. To the best of our knowledge, GR-NavSim is the largest open-vocabulary, non-template-based question-answering dataset for autonomous driving to date.
\end{itemize}

We instantiate GRAVA-8B from Qwen3-VL-8B-Instruct~\cite{bai2025qwen3vl}. Under the full NAVSIM protocol, GRAVA achieves 90.48 PDMS, improving by 10.2\% relative to its pre-RL checkpoint and achieving the best performance among the directly autoregressive driving VLAs considered in our comparison. On an internal long-tail benchmark, full GRA improves key-object compliance and Closed-loop Driving Score by 19.3\% and 20.5\% relative to action-only prediction, respectively. Controlled reasoning interventions further show that replacing low-reward GRA reasoning with higher-quality grounded reasoning directly changes the generated action and substantially improves trajectory quality. These results support the central premise of GRAVA: action-relevant physical evidence should be explicitly grounded within the reasoning and preserved through interaction analysis, decision making, and executable action generation.

\begin{figure*}[t]
  \centering
  \includegraphics[width=\textwidth]{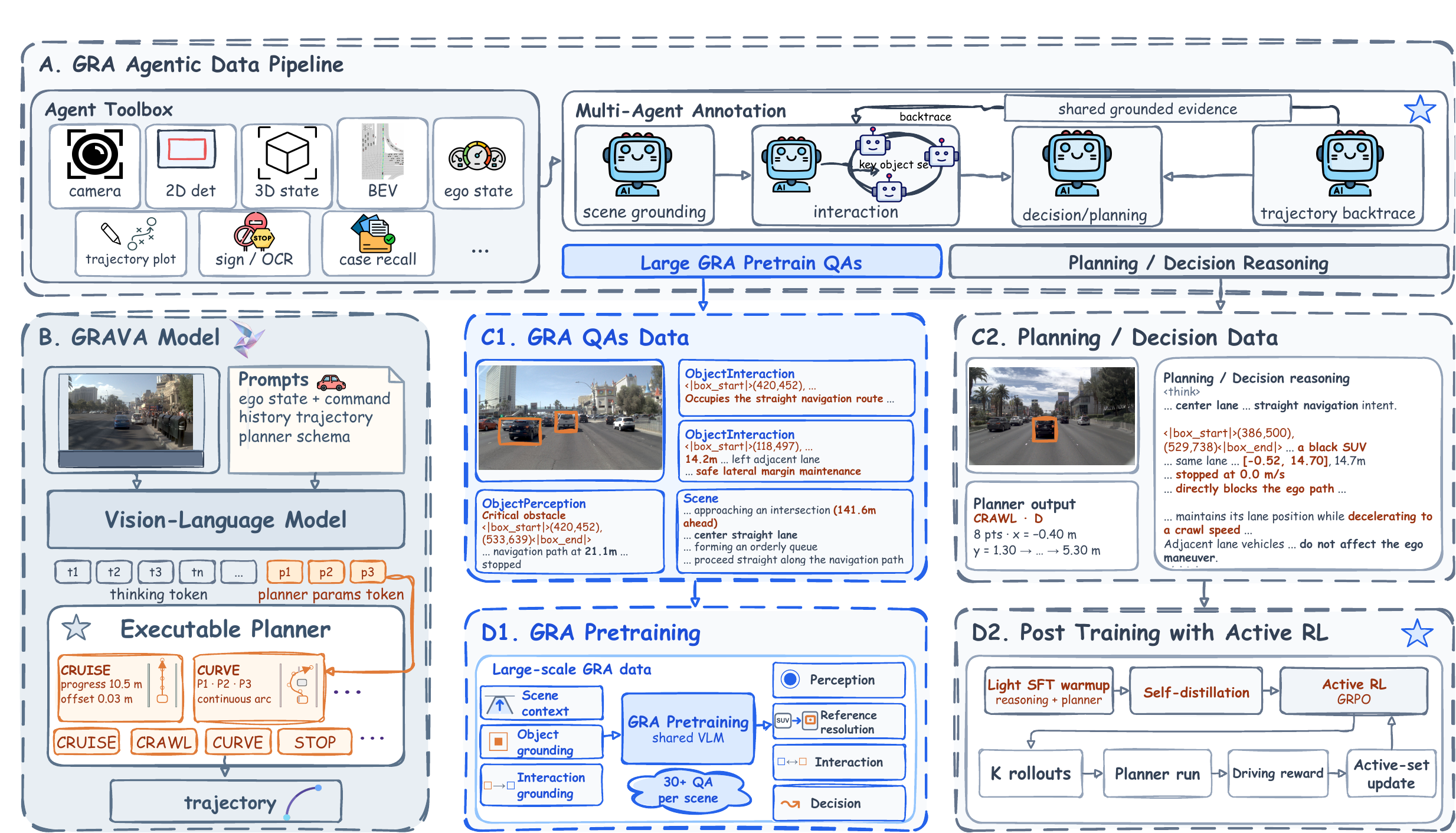}
\caption{Overview of GRAVA. (A) Agentic GRA graph construction through forward grounding and backward trajectory anchoring. (B) Autoregressive grounded reasoning and Executable Planner action generation, followed by fixed trajectory decoding. (C1--C2) Cognition and planning supervision derived from the same GRA graph. (D1--D2) Progressive training from grounded pre-training to Active RL.}
  \label{fig:overview}
\end{figure*}


\section{Related Work}

\subsection{Grounded Reasoning in Driving VLAs}

Grounded vision-language learning associates linguistic concepts with observable visual evidence. Visual Genome and GQA provide object-, relation-, and scene-graph-based supervision~\cite{krishna2016visualgenome,hudson2019gqanew}, while MDETR and Grounding DINO explicitly connect language expressions to image regions~\cite{kamath2021mdetrmodulated,liu2023groundingdino}. Region-aware multimodal models further support localized referring, question answering, and spatial reasoning~\cite{peng2024kosmos2,you2024ferret,chen2024spatialvlm}. These studies establish the visual reference resolution and spatial understanding capabilities needed to ground reasoning in scene evidence.

Driving VLMs and VLAs extend these capabilities toward reasoning about ego behavior. EMMA~\cite{hwang2025emma} jointly models scene understanding and trajectory generation, DriveLM~\cite{sima2024drivelm} organizes driving cognition through graph-structured visual question answering, and DriveVLM~\cite{tian2025drivevlm}, LMDrive~\cite{shao2024lmdrive}, and Senna~\cite{jiang2026senna} introduce language-based reasoning or high-level decisions into planning. More recent methods explicitly incorporate physical grounding and interaction structure. OpenDriveVLA~\cite{zhou2026opendrivevla} uses 2D and 3D instance-aware representations with agent--environment--ego interactions, Alpamayo-R1~\cite{nvidia2025alpamayo} constructs decision-grounded Chain-of-Causation traces, ReCogDrive~\cite{li2026recogdrive} develops hierarchical cognition before planning, and DriveAgent-R1~\cite{zheng2026driveagentr1} invokes perception tools to ground decisions in visual evidence.

These approaches strengthen the connection between scene understanding and driving reasoning. GRA focuses on preserving action-relevant evidence throughout that reasoning: visual references and physical states remain associated with the object-specific interactions and decisions they support. Its trajectory-anchored graph organizes these dependencies, and its serialization retains them within the reasoning sequence that precedes executable action generation.

\begin{figure*}[t]
    \centering
    \includegraphics[width=\textwidth]{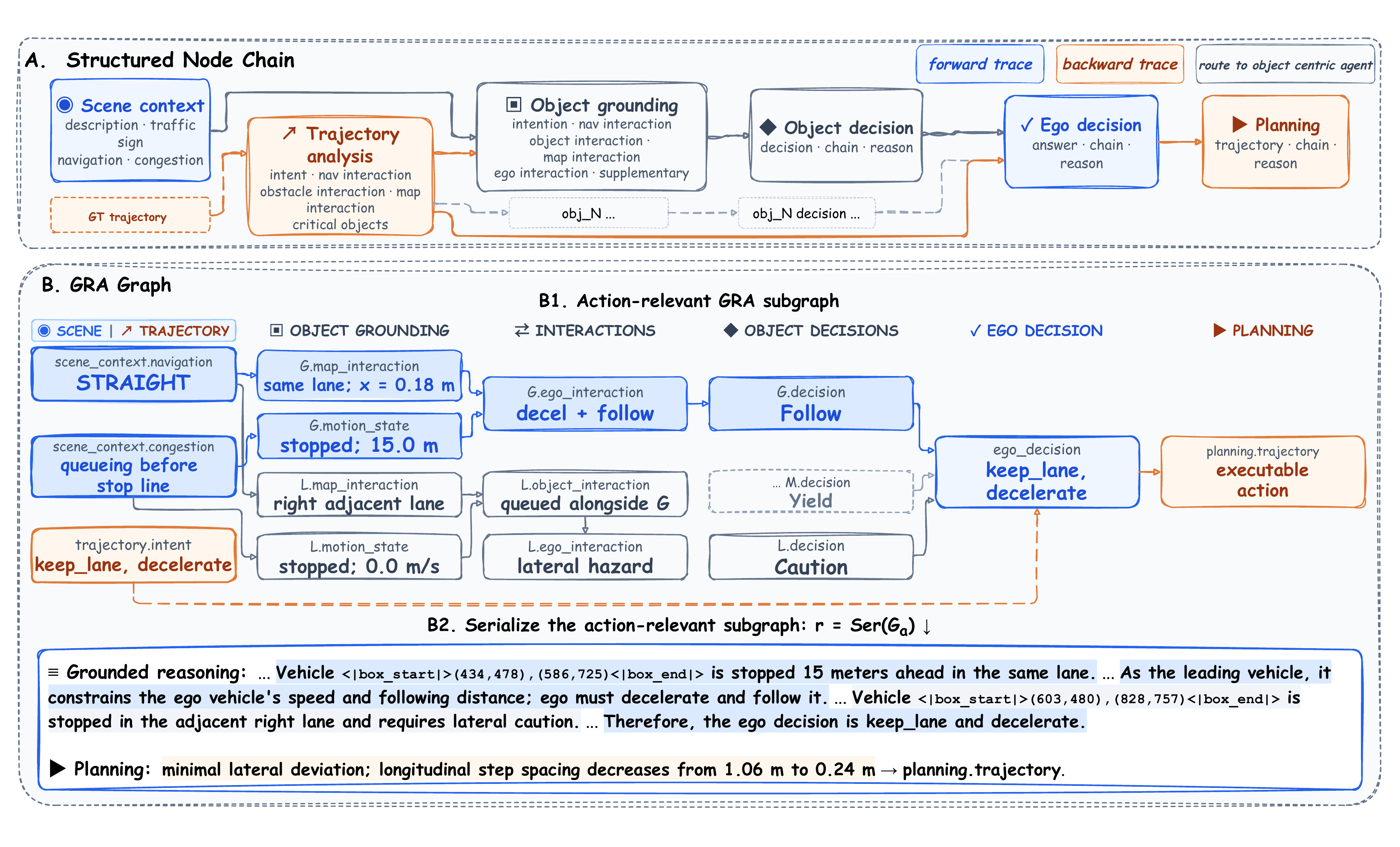}
\caption{GRA graph structure and serialization. (A) Semantic node types and their organization. (B1) Object-specific reasoning paths converge at the ego decision and connect to the action anchor. (B2) The action-relevant subgraph is serialized into grounded reasoning with visual references and physical states, followed by the planning target.}
    \label{fig:gra}
\end{figure*}

\subsection{Reasoning-to-Action Representation}

End-to-end autonomous driving has progressed from direct control and trajectory regression~\cite{codevilla2018conditional,chen2020learningbycheating,wu2022tcp} toward object-centric, bird's-eye-view, vectorized, and sparse representations that expose spatial structure to planning~\cite{li2026bev3dsurvey,hu2022stp3,renz2023plant,hu2023uniad,jiang2023vad,sun2025sparsedrive}. Diffusion and autoregressive planners further model multimodal trajectories and sequential motion~\cite{liao2025diffusiondrive,zhang2025carplanner,feng2026artemis}. For driving VLAs, the action interface must additionally connect language-based reasoning to continuous vehicle motion.

One approach retains a specialized trajectory generator alongside the VLM. DriveVLM~\cite{tian2025drivevlm} and Senna~\cite{jiang2026senna} use dedicated planning modules, ReCogDrive~\cite{li2026recogdrive} couples cognition with a diffusion planner, and DriveVLA-W0~\cite{li2026drivevlaw0} introduces a lightweight action expert after world-model learning. Qwen-Drive-1.0~\cite{zhou2026qwendrive} integrates 3D perception and driving VQA within a shared VLM while generating trajectories through an external Planning Expert. These designs provide driving-specific motion generation through a learned module distinct from the autoregressive reasoning stream.

An alternative represents agent states, trajectories, or controls directly as token sequences~\cite{philion2024trajeglish,yang2024vidar,bai2024tokenizedllm,huang2025drivegpt,pertsch2025fast}. AutoVLA~\cite{zhou2025autovla} discretizes trajectories through an action codebook, while Curious-VLA~\cite{chen2026devil} improves autoregressive trajectory generation through step-wise normalization and exploration. Related work also explicitly addresses language--action consistency: SimLingo~\cite{renz2025simlingo} aligns visual question answering with closed-loop behavior, LinkVLA~\cite{wang2026linkvla} links language and action through a shared discrete codebook, and Neuro-Symbolic Drive~\cite{gao2026neurosymbolic} derives rule-grounded reasoning traces from executable planners.

GRAVA follows the direct-action paradigm with a primitive-specific interface. The VLM generates an Executable Planner action immediately after grounded reasoning, and a fixed geometric decoder converts its parameters into continuous waypoints. The learned action choice therefore remains within $\pi_\theta(a\mid x,r)$, while geometric realization requires no additional learned trajectory policy.

\subsection{Data Construction for Grounded Driving VLAs}

Large-scale driving datasets such as BDD100K, nuScenes, Waymo Open, and nuPlan provide supervision for perception, prediction, and planning~\cite{yu2020bdd100k,caesar2020nuscenes,sun2020waymo,caesar2021nuplan}. Language-oriented datasets extend this supervision toward command grounding and visual question answering~\cite{deruyttere2019talk2car,qian2024nuscenesqa,marcu2024lingoqa}. DriveLM~\cite{sima2024drivelm} organizes driving questions through graph dependencies, while NuPlanQA~\cite{park2025nuplanqa} provides multi-view scene-understanding and ego-centric reasoning supervision. NuScenes-QA further demonstrates how scene graphs constructed from existing 3D annotations can support scalable question-answer generation. Together, these datasets establish structured sources of driving cognition supervision.

Automated and model-assisted pipelines increasingly extend such supervision to driving reasoning. Alpamayo-R1~\cite{nvidia2025alpamayo} constructs decision-grounded Chain-of-Causation traces through automatic and human-in-the-loop annotation, while ReCogDrive~\cite{li2026recogdrive} uses a hierarchical cognition-data pipeline with generation, refinement, and quality control. These approaches improve the scale and organization of reasoning data. For learning a continuous reasoning-to-action process, a central consideration is how the grounded concepts acquired through cognition supervision remain consistent with the evidence used in planning targets.

The agentic GRA data construction pipeline addresses this cross-task consistency through a shared trajectory-anchored graph. Forward grounding establishes visual references and physical states, while backward trajectory anchoring identifies the objects, interactions, and decisions relevant to the demonstrated behavior. Cognition-oriented questions and planning-oriented reasoning sequences are derived from the same graph, preserving their object identities and decision semantics. The distinction lies in using the same grounded contents both as individual cognition targets and as the evidence supporting executable action learning.

Complementary evaluation work extends trajectory-displacement measures toward closed-loop and pseudo-closed-loop driving quality~\cite{zhai2023rethinking,dauner2024navsim,jia2024bench2drive,cao2025pseudosimulation}, providing behavioral criteria alongside language-oriented supervision.

\subsection{Training Strategies for Driving VLAs}

Training driving VLAs requires adapting general-purpose vision-language capabilities to driving behavior. Supervised fine-tuning learns scene understanding, reasoning, and action prediction from annotated examples~\cite{hwang2025emma,zhou2026opendrivevla}. In imitation-centered training, the driving policy is primarily optimized to reproduce demonstrated behavior. This provides a direct route to acquiring driving skills, while its susceptibility to distribution shift motivates interactive imitation-learning methods such as DAgger~\cite{ross2011dagger}.

A broader training strategy first establishes driving cognition and then adapts it to planning. ReCogDrive~\cite{li2026recogdrive} adopts driving-specific pre-training followed by planner imitation and reinforcement learning. Qwen-Drive-1.0~\cite{zhou2026qwendrive} similarly combines perception and VQA adaptation with subsequent Planning Expert training and reward-based refinement. These staged systems distinguish the acquisition of scene-understanding capabilities from the optimization of driving behavior. Complementary self-training and reasoning-distillation methods reuse verified model-generated outputs to extend supervision beyond the original annotations~\cite{xie2020noisystudent,zelikman2022star}.

Within such training systems, RL and preference-based post-training support further policy improvement beyond supervised imitation~\cite{schulman2017ppo,rafailov2023dpo,chu2025sft}. Group-relative optimization, asymmetric policy updates, and entropy regulation provide mechanisms for learning from sampled outcomes~\cite{shao2024deepseekmath,yu2025dapo,cui2025entropy}. In driving, AlphaDrive~\cite{jiang2025alphadrive} applies group-relative optimization to reasoning, Gen-Drive~\cite{huang2025gendrive} combines reward modeling with diffusion-policy fine-tuning, and AutoVLA~\cite{zhou2025autovla} uses reinforcement fine-tuning for end-to-end VLA planning. Curious-VLA~\cite{chen2026devil} further addresses policy narrowing after imitation through iterative exploration and sample refresh.

GRAVA emphasizes grounded cognition pre-training followed by lightweight planner imitation and reinforcement-driven exploration. GRA pre-training establishes the grounded concepts used in reasoning, while planner warm-up learns their continuation into executable actions. Verified self-distillation stabilizes model-generated reasoning before Active RL explores and optimizes complete $[r;a]$ sequences according to decoded trajectory outcomes. Imitation thus establishes the initial reasoning-to-action interface, while RL refines behavior on policy-dependent recoverable scenarios where high-reward actions are reachable but not yet reliable.


\section{The GRAVA Framework}
\label{sec:method}

\subsection{Overview and GRA Formulation}
\label{sec:overview}

GRAVA generates a driving action by first identifying the relevant visual evidence and physical states, reasoning about how they constrain the ego vehicle, and then expressing the resulting decision as an executable action. Grounding is part of this reasoning process: an object's visual reference, distance, and motion state appear alongside the interactions and decisions they support. A single autoregressive VLM generates the grounded reasoning and the subsequent action, as illustrated in Fig.~\ref{fig:overview}.

\textbf{Grounded Reasoning-to-Action (GRA)} defines how this information is organized. Its graph links scene context and grounded object evidence to interactions and driving decisions; its serialization expresses those dependencies as a reasoning sequence that the VLM can learn and generate. The graph thus specifies the structure of the reasoning, while the text is its sequential realization. We first describe this generation process, then explain how graph-derived supervision and staged learning establish it.

Given visual observation $V_{\mathrm{env}}$, ego state $s_{\mathrm{ego}}$, and navigation command $c$, the model input is $x=(V_{\mathrm{env}},s_{\mathrm{ego}},c)$. Let $G_a$ denote the action-relevant GRA subgraph and $r=\operatorname{Ser}(G_a)$ its grounded reasoning sequence. Bounding-box tokens identify relevant visual regions, physical attributes describe object states, and interaction and decision statements explain their influence on ego behavior.

After generating $r$, the VLM predicts an \textbf{Executable Planner} action $a=(p,g,\phi)$, where $p$ is the motion primitive, $g$ the gear, and $\phi$ the primitive-specific parameters. The complete sequence $y=[r;a]$ is generated autoregressively with the factorization
\begin{equation}
\pi_\theta(y\mid x)=\pi_\theta(r\mid x)\pi_\theta(a\mid x,r).
\label{eq:reason_action_factorization}
\end{equation}
The two factors describe a continuous information flow. The reasoning sequence introduces grounded objects and explains their behavioral relevance; the action is then generated with that reasoning in context. GRA addresses the grounding gap (\textbf{G1}) through the content and organization of $r$, and the reasoning-to-action fragmentation gap (\textbf{G2}) by learning $a$ as its continuation within the same policy.

A fixed geometric decoder converts the action into a continuous trajectory:
\begin{equation}
\hat{\tau}=D_p\left(g,\phi;s_{\mathrm{ego}}\right).
\label{eq:trajectory_decoder}
\end{equation}
The decoder has no learnable parameters; the VLM selects the action, and $D_p$ realizes its geometry.

Training supervision is derived from a target graph constructed offline:
\begin{equation}
G^{*}=\mathcal{A}\left(x,\tau^{*},z\right),
\label{eq:target_gra}
\end{equation}
where $\tau^{*}$ is the expert trajectory, $z$ denotes privileged annotation evidence, and $\mathcal{A}$ is the agentic graph-construction process. Queries of $G^{*}$ provide cognition-oriented QA targets, while its action-relevant paths provide reasoning targets paired with expert-derived planner actions. These targets teach the model both the individual grounded concepts and their use in a complete driving decision.

Section~\ref{sec:active-rl} describes how staged learning builds this capability and then improves it using driving outcomes. The graph-construction process is used offline; at inference, GRAVA receives only $x$ and directly generates $[r;a]$. Neither an expert trajectory $\tau^{*}$ nor privileged evidence $z$ is required to produce the reasoning and action.

\subsection{Grounded Reasoning-to-Action Representation}
\label{sec:gra}

GRA organizes scene evidence according to its role in the driving decision. In the example in Fig.~\ref{fig:gra}, a vehicle is stopped 15\,m ahead in the ego lane. Its visual region identifies which vehicle is being discussed; its motion state and lane relation explain why it constrains forward motion; the resulting interaction calls for deceleration and following. A vehicle in the adjacent right lane imposes a different constraint, requiring lateral caution. GRA keeps these object-specific reasoning paths distinct before combining them into the ego decision.

The representation is \emph{trajectory anchored}: the demonstrated behavior determines which scene evidence and relations are retained. We describe its node types and dependencies, explain how the graph is constructed, and then show how its contents become grounded reasoning and training supervision.

\subsubsection{Typed Grounded Reasoning Graph}

The nodes represent semantic contents such as navigation intent, object motion state, interaction, and decision. Figure~\ref{fig:gra} makes these contents explicit through fields such as \texttt{scene\_context.navigation}, \texttt{G.motion\_state}, and \texttt{G.ego\_interaction}. Directed edges organize their dependencies from grounded evidence to object-level responses and the final ego decision. For each scene, this typed directed acyclic graph is defined as
\begin{equation}
G=(V,E,\nu,\eta),
\label{eq:gra_graph}
\end{equation}
where $V$ denotes the node set, $E\subseteq V\times V$ the directed edge set, and $\nu$ and $\eta$ specify the corresponding node and edge types. The node set is decomposed as
\begin{equation}
V=V^{S}\mathbin{\dot{\cup}}V^{O}\mathbin{\dot{\cup}}V^{\Psi}\mathbin{\dot{\cup}}V^{D}\mathbin{\dot{\cup}}\{v^{\delta},v^{a}\},
\label{eq:gra_nodes}
\end{equation}
where $V^{S}$ contains scene- and route-context nodes, $V^{O}$ contains grounded object nodes, $V^{\Psi}$ contains interaction nodes, and $V^{D}$ contains object-level decision nodes. The ego-level decision node $v^{\delta}$ aggregates the responses induced by individual action-relevant objects into the final driving decision, while $v^{a}$ serves as the terminal action anchor. The action anchor marks the endpoint of the grounded reasoning structure but does not itself contain the parameters of the Executable Planner.

The grounding information associated with an object connects its visual reference to the physical states used in subsequent reasoning. We collect this information in the object node $o_i$, writing, for each action-relevant object $i$,
\begin{equation}
o_i=(m_i,b_i,\sigma_i),
\label{eq:gra_object}
\end{equation}
where $m_i$ denotes its linguistic reference, $b_i\in[0,1000)^4$ is its normalized 2D bounding box, and $\sigma_i$ denotes its ego-centric physical state. The state $\sigma_i$ contains the semantic category and attributes such as relative position, distance, speed, and heading. The bounding box resolves which visible object a statement concerns; its physical state provides the information needed to reason about the object's effect on driving. Both remain associated with the object when it appears in later interaction and decision statements.

Let $\mathcal{C}$ denote the set of critical objects whose states or interactions constrain the demonstrated ego behavior. For each $i\in\mathcal{C}$, object-specific reasoning proceeds from scene context and grounded object evidence through one or more interaction nodes to an object-level decision:
\begin{equation}
(V^{S},o_i)\longrightarrow \{\psi_i^{\ell}\}_{\ell}\longrightarrow d_i,\qquad i\in\mathcal{C}.
\label{eq:object_reasoning_path}
\end{equation}
The resulting object-level decisions are subsequently combined at the ego level:
\begin{equation}
(V^{S},\{d_i\}_{i\in\mathcal{C}})\longrightarrow v^{\delta}\longrightarrow v^{a}.
\label{eq:ego_reasoning_path}
\end{equation}
In the running example, the lead-vehicle branch contributes the need to decelerate and follow, while the adjacent-vehicle branch contributes lateral caution. Their combination supports the ego decision to keep the lane and decelerate. The branch-and-merge topology records which object supports each response before the responses are combined.

To prevent reasoning statements from becoming detached from observable evidence, GRA imposes structural grounding constraints. Every retained interaction or object-level decision must be traceable to at least one grounded object:
\begin{equation}
\forall v\in V^{\Psi}\cup V^{D},\qquad \mathrm{Anc}_{G}(v)\cap V^{O}\neq\emptyset,
\label{eq:grounded_ancestor}
\end{equation}
where $\mathrm{Anc}_{G}(v)$ denotes the ancestor set of node $v$ in $G$. Each object-level decision must also contribute to the ego-level decision:
\begin{equation}
\forall d_i\in V^{D},\qquad d_i\longrightarrow v^{\delta},
\label{eq:decision_reachability}
\end{equation}
and the ego-level decision must ultimately reach the action anchor:
\begin{equation}
v^{\delta}\longrightarrow v^{a}.
\label{eq:action_reachability}
\end{equation}
where $\longrightarrow$ denotes directed path reachability. These constraints connect each retained interaction or object-level decision to grounded object evidence and link the object-level decisions to the final ego action anchor.

\begin{table}[t]
\caption{Executable Planner action schema and primitive-specific parameter spaces $\Phi_p$ in the ego frame.}
\label{tab:planner_schema}
\centering
\small
\setlength{\tabcolsep}{2.5pt}
\renewcommand{\arraystretch}{1.05}
\begin{tabular}{c c c p{0.32\columnwidth}}
\hline
Primitive $p$ & Parameters $\phi\in\Phi_p$ & Shape & Motion regime \\
\hline
STOP & endpoint & $\mathbb{R}^{2}$ & near-stationary or stopping motion \\
CRAWL & points & $\mathbb{R}^{8\times2}$ & low-speed stop-and-go \\
CURVE & $p_1,p_2,p_3$ & $\mathbb{R}^{3\times2}$ & turning and high-curvature motion \\
CRUISE & \begin{tabular}[c]{@{}c@{}}progress\_end,\\ $v_{\mathrm{end}}$, lat\_end\end{tabular} & $\mathbb{R}^{3}$ & lane keeping and car following \\
\hline
\end{tabular}
\end{table}

\subsubsection{Trajectory-Anchored Graph Construction}

The offline construction process in Eq.~\eqref{eq:target_gra} combines two directions, illustrated in Fig.~\ref{fig:overview}. \emph{Forward scene grounding} establishes what is present: it identifies candidate traffic participants and scene elements and associates them with visual regions and physical states. This produces the evidence represented by $V^{S}$ and $V^{O}$, but does not by itself determine which objects constrain the demonstrated action.

\emph{Backward trajectory anchoring} establishes behavioral relevance: starting from the expert trajectory $\tau^{*}$, it traces the demonstrated behavior back to the critical objects, interactions, and decisions that support it. In the running example, the lead vehicle's stopped state and same-lane position provide the evidence for the deceleration-and-following response. Linking that response back to the grounded object makes its role in the demonstrated behavior explicit.

Together, the two directions connect observable scene evidence with action-relevant reasoning. Objects and relations that do not contribute to the demonstrated decision are excluded from the action-relevant structure. The resulting $G^{*}$ records the grounded states, interactions, and decisions from which the learning targets are derived. Privileged evidence is used only in offline construction and training.

\subsubsection{Action-Relevant Graph Serialization}

Serialization makes the graph's reasoning structure available to an autoregressive VLM. It expresses grounded states, interactions, and decisions in their dependency order, so that the model learns how physical evidence leads to a driving response. Let $\operatorname{Anc}_{G}(v^{a})$ denote the strict ancestor set of the action anchor. We extract the action-relevant reasoning subgraph as
\begin{equation}
G_a=G[\operatorname{Anc}_{G}(v^{a})],
\label{eq:action_subgraph}
\end{equation}
where $G[\cdot]$ denotes the induced subgraph. The corresponding grounded reasoning sequence is obtained as
\begin{equation}
r=\operatorname{Ser}(G_a).
\label{eq:gra_action_serialization}
\end{equation}

Serialization follows the topological structure of $G_a$. Grounded objects are introduced with their linguistic references, normalized bounding-box tokens, and physical states. Subsequent statements retain the object references while describing interactions and object-level decisions, before converging at the ego decision.

Figure~\ref{fig:gra}(B2) illustrates this transition for the running example. The text first identifies the lead vehicle by its bounding box and states that it is stopped 15\,m ahead in the same lane. It then explains how that vehicle constrains speed and following distance, retains the adjacent vehicle's lateral constraint, and concludes with lane keeping and deceleration. The graph nodes thus become connected reasoning statements, with the visual references and physical quantities embedded in the reasoning itself.

The action anchor itself is not included in the reasoning sequence $r$. Instead, as defined in Sec.~\ref{sec:overview}, the Executable Planner action $a$ is generated immediately after $r$. The complete progression is therefore
\begin{equation}
G_a\longrightarrow r\longrightarrow a\longrightarrow \hat{\tau}.
\label{eq:gra_to_action}
\end{equation}
This progression describes how the graph organizes the target reasoning and how that reasoning leads into action generation. At inference, the VLM directly generates the sequence: it introduces the grounded evidence, reasons about its consequences, and continues with the planner action. Section~\ref{sec:planner} details the action representation and its trajectory decoder.

\begin{figure*}[t]
  \centering
  \makebox[\textwidth][c]{%
    \includegraphics[width=\textwidth]{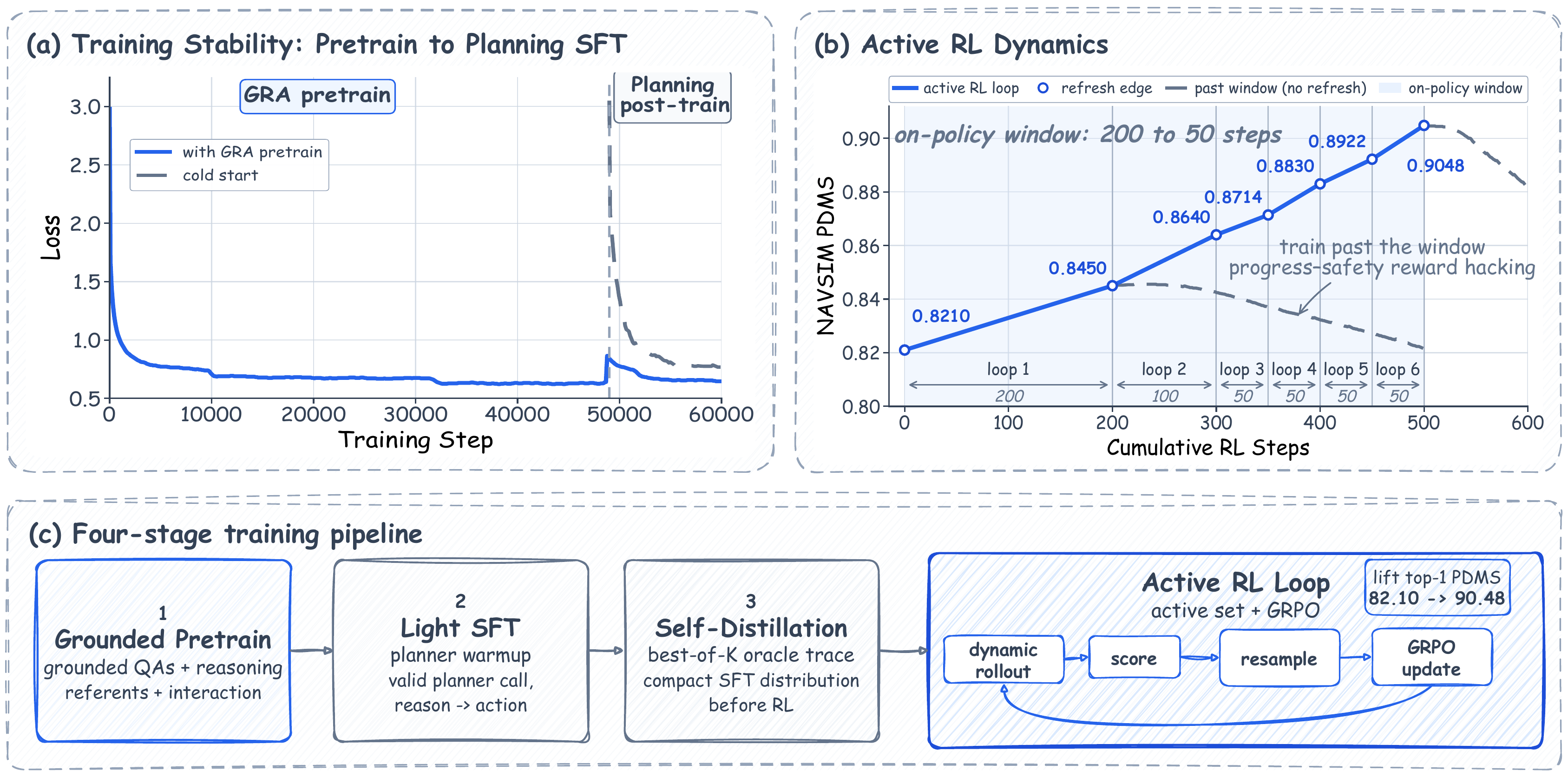}%
  }
\caption{Progressive training and Active RL dynamics of GRAVA. (a) Training loss with GRA pre-training and with planner training from a cold start (dashed). (b) Greedy NAVSIM PDMS across Active RL loops; shaded regions mark on-policy optimization windows. (c) Grounded pre-training, planner warm-up, verified self-distillation, and Active RL.}
  \label{fig:training}
\end{figure*}

\subsubsection{Shared Cognition and Planning Supervision}

The target graph supports two complementary forms of supervision. Cognition-oriented questions teach the model to identify and reason about individual parts of the structure; planning-oriented sequences teach it to use those contents together to produce an action. For cognition pre-training, graph queries sample a question $q$ and answer $u$ concerning grounded objects, physical states, interaction relations, or driving decisions. The corresponding objective is
\begin{equation}
\mathcal{L}_{\mathrm{QA}}=-\mathbb{E}_{(x,q,u)}\left[\log\pi_{\theta}(u\mid x,q)\right].
\label{eq:qa_loss}
\end{equation}

For planning-oriented learning, the action-relevant portion of the same target graph is serialized as
\begin{equation}
r^{*}=\operatorname{Ser}(G_a^{*}),
\label{eq:target_reasoning}
\end{equation}
and paired with the corresponding target Executable Planner action $a^{*}$. The planning objective is
\begin{equation}
\mathcal{L}_{\mathrm{plan}}=-\mathbb{E}_{(x,r^{*},a^{*})}\left[\log\pi_{\theta}\left([r^{*};a^{*}]\mid x\right)\right].
\label{eq:planning_loss}
\end{equation}

Both objectives use the object identities, physical states, interactions, and decisions recorded in $G^{*}$. Grounding learned through a local question therefore appears again as part of the reasoning that supports an action. Shared supervision connects the acquisition of individual cognition capabilities to their use in the full reasoning-to-action sequence.

\subsection{Executable Planner}
\label{sec:planner}

Grounded reasoning explains how the scene constrains the ego behavior; the next step is to express that behavior as vehicle motion. The \textbf{Executable Planner} provides the action vocabulary for this continuation. The VLM selects a motion primitive and its parameters immediately after generating $r$, and a fixed geometric decoder converts this structured action into a trajectory. This keeps the transition from reasoning to action within the same autoregressive policy while allowing different motion regimes to use different geometric parameterizations.

The compact action is $a=(p,g,\phi)$. The primitive set $\mathcal{P}$ comprises STOP, CRAWL, CURVE, and CRUISE, with gear $g\in\mathcal{M}=\{\mathrm{D},\mathrm{R}\}$ and parameters $\phi\in\Phi_p$. The executable action space is
\begin{equation}
\mathcal{A}_{\mathrm{exec}}=\bigcup_{p\in\mathcal{P}}\left(\{p\}\times\mathcal{M}\times\Phi_p\right).
\label{eq:exec_action_space}
\end{equation}
Table~\ref{tab:planner_schema} summarizes the primitive-specific parameterizations. STOP represents near-stationary or stopping motion; CRAWL retains waypoint-level flexibility for irregular low-speed stop-and-go motion; CURVE describes turning and high-curvature trajectories; and CRUISE expresses lane-following and car-following motion through terminal motion variables.

The action is serialized as a continuation of the grounded reasoning:
\begin{equation}
\operatorname{Ser}(a)=\{\texttt{planner}:p,\,\texttt{gear}:g,\,\texttt{params}:\phi\}.
\label{eq:planner_action_serialization}
\end{equation}
The fixed decoder in Eq.~\eqref{eq:trajectory_decoder} converts a valid action into an eight-waypoint trajectory $\hat{\tau}\in\mathbb{R}^{8\times2}$ over a 4-s horizon. Because the decoder has no learnable parameters, action selection remains in $\pi_\theta(a\mid x,r)$, with the grounded evidence and decisions in context. The Executable Planner is an action representation, not an additional learned planning policy.

\subsubsection{Supervised Action Target Construction}

During supervised learning, the target planner action $a^{*}=(p^{*},g^{*},\phi^{*})$ is deterministically derived from the expert trajectory $\tau^{*}$. The target encoder first assigns the motion primitive $p^{*}$ according to the expert motion profile and navigation command, determines the driving gear $g^{*}$ from the motion direction, and then fits only the parameters associated with the corresponding parameter space $\Phi_{p^{*}}$. The resulting parameters $\phi^{*}$ are quantized and serialized together with $p^{*}$ and $g^{*}$ to form the action tokens appended to the target grounded reasoning sequence $r^{*}$.

This target construction preserves the flexibility of each motion regime: CRAWL retains eight waypoints for irregular low-speed motion, while STOP, CURVE, and CRUISE use lower-dimensional parameters. Combined with the graph-derived reasoning, it yields the supervised sequence $y^{*}=[r^{*};a^{*}]$, which teaches action prediction as a continuation of grounded reasoning.

\subsection{Learning Grounded Reasoning-to-Action}
\label{sec:active-rl}

GRAVA learns the reasoning-to-action process in stages, as illustrated in Fig.~\ref{fig:training}. GRA pre-training develops grounded cognition from graph-derived questions. Planner warm-up connects that cognition to executable actions through complete reasoning-to-action targets. Verified self-distillation then incorporates valid model-generated reasoning, and \textbf{Active RL} optimizes the resulting sequences using the driving outcomes of their decoded actions.

\paragraph{Grounded pre-training and planner warm-up}
GRA pre-training uses the cognition objective in Eq.~\eqref{eq:qa_loss} to learn scene understanding, reference resolution, physical-state reasoning, interactions, and decision semantics. Planner warm-up then uses the sequence objective in Eq.~\eqref{eq:planning_loss} to learn how these contents support an executable action. Because both forms of supervision are derived from GRA, the grounded concepts learned through individual questions reappear in the reasoning that precedes the planner output.

Supervision establishes this mapping, but driving outcomes provide a further criterion for choosing among model-generated sequences. Different actions can be geometrically plausible yet differ in safety, progress, and interaction behavior. The following stages first retain verified model-generated reasoning and then refine the complete $[r;a]$ sequence according to the quality of its decoded trajectory.

\paragraph{Verified self-distillation}
Starting from the planner warm-up checkpoint, GRAVA samples complete reasoning-to-action sequences $y_i=[r_i;a_i]$. Candidates are selected according to the closed-loop driving rewards of their decoded trajectories. For each retained sample, the distilled target is defined as
\begin{equation}
y_i^{+}=[r_i^{+};a_i^{+}],
\label{eq:sd-target}
\end{equation}
where $r_i^{+}$ and $a_i^{+}$ denote the reasoning and Executable Planner action from the same reward-selected rollout. The resulting checkpoint initializes Active RL and serves as its frozen reference policy.

\paragraph{Active RL: selecting recoverable scenarios}
Active RL focuses on scenes where the current model can discover a good action but does not yet produce it reliably. Such \emph{recoverable} scenarios provide contrasting reasoning-to-action sequences for the same observation: some lead to high-reward behavior, while others lead to substantially worse outcomes. For each input $x$, the current policy generates $K$ complete rollouts $\{y_i=[r_i;a_i]\}_{i=1}^{K}$. Each valid planner action $a_i=(p_i,g_i,\phi_i)$ is deterministically decoded and evaluated using the driving reward
\begin{equation}
s_i=
\mathcal{R}\!\left(
D_{p_i}(g_i,\phi_i;s_{\mathrm{ego}});x
\right).
\label{eq:rollout-score}
\end{equation}

We first identify scenarios that remain insufficiently solved under the policy's greedy output:
\begin{equation}
\mathcal{D}_{g}(\theta)
=
\left\{
x:s_{\mathrm{greedy}}(x)<\tau_g
\right\}.
\label{eq:greedy-candidate-set}
\end{equation}
Within this candidate pool, a scenario is retained only when its sampled rollouts contain a sufficiently good solution, a substantially worse alternative, and meaningful reward variation:
\begin{equation}
\mathcal{D}_{\mathrm{act}}(\theta)
=
\left\{
x\in\mathcal{D}_{g}(\theta)
\,\middle|\,
\begin{array}{l}
\max_i s_i\geq\tau_{\mathrm{hi}},\\[-1pt]
\min_i s_i<\tau_{\mathrm{lo}},\\[-1pt]
\operatorname{std}_i(s_i)\geq\tau_{\sigma}
\end{array}
\right\}.
\label{eq:recoverable-gap}
\end{equation}
The high-reward criterion selects scenes where improved behavior is reachable; the low-reward and variance criteria retain scenes where sampled outcomes still differ substantially. Together with the greedy filter, they identify where a successful sampled behavior has yet to become a reliable policy output. All candidate rollouts are generated by GRAVA itself, so this selection requires no additional action teacher.

\paragraph{Outcome-based reasoning-to-action optimization}
For NAVSIM, $\mathcal{R}$ is defined as the PDMS of the trajectory decoded from each generated Executable Planner action. Outputs that cannot be parsed or violate the predefined executable schema receive zero reward. Active RL therefore evaluates generated actions according to their final driving outcomes rather than supervising motion primitives or planner parameters with manually designed intermediate rewards.

Because every rollout contains both grounded reasoning $r_i$ and executable action $a_i$, the trajectory-level reward provides outcome feedback for the complete sequence $y_i=[r_i;a_i]$. Let $\bar{s}$ and $\sigma_s$ denote the mean and standard deviation of the rewards within a rollout group. The normalized group-relative advantage is
\begin{equation}
\hat{A}_i
=
\frac{s_i-\bar{s}}
{\sigma_s+\epsilon},
\label{eq:normalized-advantage}
\end{equation}
where $\epsilon>0$ provides numerical stability. For token $y_{i,t}$, the importance ratio between the updated policy and the frozen behavior policy that generated the rollout is
\begin{equation}
\rho_{i,t}(\theta)
=
\frac{
\pi_{\theta}(y_{i,t}\mid x,y_{i,<t})
}{
\pi_{\theta_{\mathrm{old}}}(y_{i,t}\mid x,y_{i,<t})
}.
\label{eq:importance-ratio}
\end{equation}

Following DAPO-style asymmetric clipping~\cite{shao2024deepseekmath,yu2025dapo}, the token-level objective is
\begin{equation}
\begin{aligned}
\ell_{i,t}(\theta)
=
\min\Big\{
&\rho_{i,t}(\theta)\hat{A}_i,\\
&
\operatorname{clip}\!\left(
\rho_{i,t}(\theta),
1-\epsilon_{-},
1+\epsilon_{+}
\right)\hat{A}_i
\Big\}.
\end{aligned}
\label{eq:clipped-objective}
\end{equation}
The sequence-level Active RL objective is
\begin{equation}
\begin{aligned}
\mathcal{J}_{\mathrm{RL}}(\theta)
=
\mathbb{E}_{\substack{
x\sim\mathcal{D}_{\mathrm{act}}(\theta_{\mathrm{old}})\\
y_{1:K}\sim\pi_{\theta_{\mathrm{old}}}(\cdot\mid x)
}}
\Bigg[
&\frac{1}{K}
\sum_{i=1}^{K}
\frac{1}{|y_i|}
\sum_{t=1}^{|y_i|}
\ell_{i,t}(\theta)\\
&-\beta
D_{\mathrm{KL}}
\left(
\pi_{\theta}\Vert\pi_{\mathrm{ref}}
\right)
\Bigg].
\end{aligned}
\label{eq:rl-objective}
\end{equation}
Here, $\pi_{\mathrm{ref}}$ is the frozen checkpoint obtained after verified self-distillation, $\beta$ controls the KL regularization strength, and $\epsilon_{-}$ and $\epsilon_{+}$ denote the asymmetric clipping bounds. The same rollout-level advantage $\hat{A}_i$ updates the token probabilities of both $r_i$ and $a_i$. Driving feedback thus optimizes the grounded reasoning together with the action generated from it, retaining the complete sequence as the learning unit.

After each optimization window, the updated checkpoint is evaluated under greedy decoding and used to reconstruct the active set for the next loop. As the policy improves, scenarios that become reliably solvable are removed from the active set, while scenarios that remain recoverable but unstable determine the subsequent optimization distribution. The iterative process is summarized as
\begin{equation}
\pi_{\theta^{(k)}}
\rightarrow
\mathcal{D}_{\mathrm{act}}^{(k)}
\rightarrow
\pi_{\theta^{(k+1)}}.
\label{eq:active-rl-loop}
\end{equation}
This alternation forms the Active RL Loop: trajectory rewards refine complete reasoning-to-action sequences, and renewed sampling identifies the next scenes on which to train. Across the learning stages, the model progresses from understanding individual grounded concepts to using them in action generation and refining that generation through driving outcomes.


\section{Data and Evaluation Protocols}
\label{sec:data_eval}

\begin{figure}[!t]
\centering
\includegraphics[width=\linewidth,trim=7.5bp 7.3bp 6.5bp 6.5bp,clip]{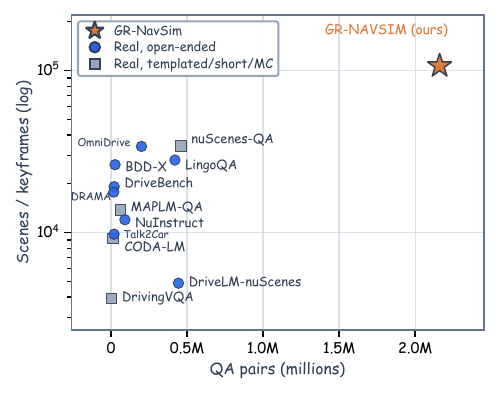}
\caption{Scale comparison of GR-NavSim and representative public driving QA datasets by QA count and source scenes or keyframes. The vertical axis is logarithmic; marker shapes distinguish answer formats.}
\label{fig:dataset-scale}
\end{figure}

\begin{table}[t]
\centering
\footnotesize
\setlength{\tabcolsep}{3pt}
\renewcommand{\arraystretch}{1.15}
\caption{Composition of the GR-NavSim training corpus.}
\label{tab:corpus}

\begin{tabularx}{\linewidth}{
@{}
>{\raggedright\arraybackslash}p{0.40\linewidth}
>{\centering\arraybackslash}p{0.13\linewidth}
>{\raggedright\arraybackslash}X
@{}
}
\toprule
Corpus component & Count & Role \\
\midrule
Driving scenes with grounded state labels
& 107K
& Grounded objects, physical states, interactions, and decisions \\

Grounded question-answer pairs
& 2.2M
& Grounded scene understanding and interaction reasoning \\

Front-camera Executable Planner examples
& 415K
& Planner warm-up and self-distillation \\

Grounded trajectory traces
& 70K
& GRA paths paired with planner actions \\
\bottomrule
\end{tabularx}

\end{table}

\subsection{NAVSIM and GR-NavSim}
\label{sec:navsim-protocol}

NAVSIM provides real-world driving scenes and a held-out planning benchmark~\cite{dauner2024navsim}. We construct \textbf{GR-NavSim} from its training scenes using the GRA-centered pipeline, obtaining 107K annotated scenes and 2.2M grounded question-answer pairs, together with Executable Planner examples and grounded trajectory traces. Cognition and planning targets share the same graph-derived object identities, physical states, interactions, and decisions. Table~\ref{tab:corpus} summarizes the corpus, and Fig.~\ref{fig:dataset-scale} compares its scale with public driving QA datasets. Planning is evaluated on NAVSIM navtest under the official full protocol using the PDM Score (PDMS) and one greedy autoregressive completion per scene.

\subsection{Internal Long-Tail Benchmark}
\label{sec:internal-benchmark}

Existing public driving benchmarks offer limited coverage of complex long-tail interactions. Models can achieve high scores primarily by fitting expert trajectories, making it difficult to fully assess the benefits of complex reasoning and reinforcement-driven exploration.

To evaluate these capabilities, we construct an internal long-tail benchmark covering route obstructions, lane borrowing, and road hazards. These scenarios require the ego vehicle to determine when to pass, when to wait, and how to plan a safe, feasible trajectory through the scene, accounting for obstacle states, available space, and interaction constraints. The benchmark contains 50K clips and 700K frames.

Each annotated key object is represented as $o_j=(t_j,\Omega_j,p_j)$, where $t_j$ is its semantic type, $\Omega_j$ its manually annotated BEV region, and $p_j\in\{\mathrm{passable},\mathrm{non\mbox{-}passable}\}$ its passability. This representation covers physical obstacles, construction regions, and road-surface hazards.

\paragraph{Closed-loop Driving Score}
The logged future trajectory is not necessarily the only valid solution for a scene, as it may reflect conservative waiting behavior or characteristics of the original data-collection policy. We therefore roll out each predicted trajectory and evaluate its safety, Key-Object Compliance (KOC), route progress, and comfort. Safety is a hard gate, $\mathrm{Safety}=G_{\mathrm{col}}(\tau)G_{\mathrm{nds}}(\tau)$: the Collision Gate equals one for rollouts without at-fault collisions, and the Non-Drivable-Space Gate equals one when the ego footprint remains within drivable space. The aggregate score is
\begin{equation}
\mathrm{CDS}=\mathrm{Safety}\left(0.45\,\mathrm{KOC}+0.35\,\mathrm{Progress}+0.20\,\mathrm{Comfort}\right).
\label{eq:cds}
\end{equation}
Progress is the maximum route displacement of the ego front edge, normalized by the maximum reachable distance and clipped to $[0,1]$; Comfort measures dynamic feasibility and smoothness. The larger KOC weight emphasizes correct responses to interaction-critical objects.

\paragraph{Key-Object Compliance}
A pass-or-hold annotation specifies the desired interaction decision but does not by itself determine whether a generated trajectory actually executes that decision. KOC therefore converts each annotated key-object label into a geometric constraint on the simulated trajectory. This formulation allows multiple geometrically different trajectories to satisfy the same interaction constraint rather than requiring agreement with a single logged future trajectory.

We project each key-object region onto the route as $I_j=[s_j^{-},s_j^{+}]$. Only objects intersecting the route interaction corridor within the evaluation horizon are retained; unreachable pass-required objects are excluded. The applicable pass and hold constraints are
\begin{equation}
\begin{aligned}
c_j^{\mathrm{pass}}&=\mathbb{I}\!\left[s_{\mathrm{ego,front}}^{\max}>s_j^{-}+d_{\mathrm{trigger}}\right],\\
U&=\max\left(s_{j_0}^{-}-d_{\mathrm{follow}},U_{\min}\right),\\
c^{\mathrm{hold}}&=\mathbb{I}\!\left[s_{\mathrm{ego,front}}^{\max}\leq U\right],
\end{aligned}
\label{eq:pass_hold_constraints}
\end{equation}
where $s_{\mathrm{ego,front}}^{\max}$ is the maximum route progress of the ego front edge and $j_0$ is the nearest hold-required object. The margins $d_{\mathrm{trigger}}$, $d_{\mathrm{follow}}$, and $U_{\min}$ are fixed across models. Only pass-required objects before the nearest hold boundary contribute, allowing a trajectory to pass an earlier object and stop behind a later one. For the applicable constraint set $\mathcal{K}$,
\begin{equation}
\mathrm{KOC}=\frac{1}{|\mathcal{K}|}\sum_{c\in\mathcal{K}}c.
\label{eq:koc}
\end{equation}
Progress is set to zero if the ego vehicle crosses a hold boundary. Samples without applicable key-object constraints are excluded from KOC aggregation, and all models use the same relevance thresholds and invalid-sample filtering rules.

\begin{table*}[!t]
\centering
\footnotesize
\setlength{\tabcolsep}{3.2pt}
\caption{Single-sample performance comparison on NAVSIM navtest v1, grouped by model design.}
\label{tab:navsim-context}
\begin{tabular}{@{}l l c r r r r r r@{}}
\toprule
Method & Backbone & Views & NC & DAC & EP & TTC & Comf & PDMS $\uparrow$ \\
\midrule
\multicolumn{9}{l}{\textit{Non-VLA methods}} \\
UniAD \cite{hu2023uniad} & \textemdash & Cam & 97.8 & 91.9 & 78.8 & 92.9 & 100.0 & 83.4 \\
PARA-Drive \cite{weng2024paradrive} & \textemdash & Cam & 97.9 & 92.4 & 79.3 & 93.0 & 99.8 & 84.0 \\
TransFuser \cite{chitta2023transfuser} & \textemdash & C+L & 97.7 & 92.8 & 79.2 & 92.8 & 100.0 & 84.0 \\
DiffusionDrive \cite{liao2025diffusiondrive} & \textemdash & C+L & 98.2 & 96.2 & 82.2 & 94.7 & 100.0 & 88.1 \\
\addlinespace
\multicolumn{9}{l}{\textit{VLAs with additional trajectory head or world model}} \\
DriveVLA-W0 \cite{li2026drivevlaw0} & Emu-3-8B & Cam & 98.7 & 99.1 & 87.6 & 97.1 & 100.0 & 90.2 \\
ExploreVLA \cite{sheng2026explorevla} & \textemdash & Cam & 98.8 & 98.4 & 83.5 & 96.5 & 99.9 & 90.4 \\
ReCogDrive \cite{li2026recogdrive} & InternVL2-8B & Cam & 98.2 & 97.8 & 83.5 & 95.2 & 100.0 & 89.6 \\
\addlinespace
\multicolumn{9}{l}{\textit{VLAs with direct action generation}} \\
AutoVLA one-shot \cite{zhou2025autovla} & Qwen2.5-VL-3B & Cam & 96.9 & 92.4 & 75.8 & 88.1 & 99.9 & 80.5 \\
AutoVLA Post-RFT \cite{zhou2025autovla} & Qwen2.5-VL-3B & Cam & 98.4 & 95.6 & 81.9 & 98.0 & 99.9 & 89.1 \\
Curious-VLA \cite{chen2026devil} & Qwen2.5-VL-3B & Cam & 98.4 & 96.9 & 88.5 & 97.9 & 98.1 & 90.3 \\
AdaThinkDrive \cite{luo2025adathinkdrive} & InternVL3-8B & Cam & 98.4 & 97.8 & 84.4 & 95.2 & 100.0 & 90.3 \\
GRAVA before Active RL & Qwen3-VL-8B & Cam & 96.1 & 91.2 & 78.8 & 91.0 & 99.7 & 82.1 \\
GRAVA after Active RL & Qwen3-VL-8B & Cam & 98.8 & 97.6 & 83.5 & 97.1 & 100.0 & 90.5 \\
\bottomrule
\end{tabular}
\par\smallskip
\begin{minipage}{\linewidth}
\footnotesize\raggedright
GRAVA is evaluated before and after Active RL under the same inference protocol. Cam and C+L denote camera-only and camera-plus-LiDAR inputs. NC: no-at-fault collision; DAC: drivable-area compliance; EP: ego progress; TTC: time to collision; Comf: comfort; PDMS: PDM Score. Higher is better for all metrics.
\end{minipage}

\par\bigskip
\centering
\footnotesize
\setlength{\tabcolsep}{2.8pt}
\caption{Component ablation on NAVSIM navtest under the full single-sample evaluation protocol.}
\label{tab:navsim-ablation}

\begin{tabular}{@{}l l l c c r r r r r r@{}}
\toprule
Pre-training & Reasoning condition & Action representation & SD & Active RL & NC & DAC & EP & TTC & Comf & PDMS $\uparrow$ \\
\midrule
None & Full GRA & Executable Planner & \checkmark & \checkmark & 95.3 & 90.0 & 75.4 & 90.9 & 99.8 & 79.80 \\
GRA & Full GRA & Direct waypoints & \checkmark & \checkmark & 97.7 & 95.0 & 82.4 & 94.2 & 99.9 & 87.23 \\
GRA & Full GRA & Executable Planner & \textemdash & \textemdash & 96.0 & 91.3 & 77.3 & 91.7 & 99.9 & 81.74 \\
GRA & Full GRA & Executable Planner & \checkmark & \textemdash & 96.1 & 91.2 & 78.8 & 91.0 & 99.7 & 82.10 \\
GRA & Full GRA & Executable Planner & \checkmark & \checkmark & 98.8 & 97.6 & 83.5 & 97.1 & 100.0 & 90.48 \\
\bottomrule
\end{tabular}
\par\smallskip
\begin{minipage}{\linewidth}
\footnotesize\raggedright
Each scene uses one greedy completion. SD denotes verified self-distillation. Metric abbreviations follow Table~\ref{tab:navsim-context}; higher is better for all metrics.
\end{minipage}

\par\bigskip
\centering
\footnotesize
\setlength{\tabcolsep}{4.0pt}
\caption{Ablation of pre-training, reasoning structure, and Active RL on the internal long-tail benchmark.}
\label{tab:internal-result}

\begin{tabular}{@{}l l c r r r r r r@{}}
\toprule
Pre-training & Reasoning condition & Active RL & Col. & NDS & KOC & Progress & Comfort & CDS $\uparrow$ \\
\midrule
None & Full GRA & \checkmark & 95.8 & 94.7 & 81.7 & 89.6 & 98.8 & 79.7 \\
Ungrounded & Full GRA & \checkmark & 96.3 & 95.4 & 84.2 & 90.7 & 99.1 & 82.2 \\
GRA & Action only & \checkmark & 94.6 & 93.2 & 77.4 & 86.9 & 97.9 & 74.8 \\
GRA & Coarse reasoning & \checkmark & 95.2 & 94.1 & 80.6 & 88.4 & 98.6 & 77.9 \\
GRA & Grounded objects only & \checkmark & 97.1 & 96.4 & 87.9 & 91.3 & 99.3 & 85.5 \\
\midrule
GRA & Full GRA & \textemdash & 93.1 & 94.9 & 69.8 & 90.1 & 98.9 & 73.1 \\
GRA & Full GRA & \checkmark & 98.2 & 97.6 & 92.3 & 92.9 & 99.6 & 90.1 \\
\bottomrule
\end{tabular}
\par\smallskip
\begin{minipage}{\linewidth}
\footnotesize\raggedright
All variants use the Executable Planner and verified self-distillation. Col. and NDS denote the pass rates of the Collision Gate and Non-Drivable-Space Gate. KOC: Key-Object Compliance; CDS: Closed-loop Driving Score. Higher is better for all metrics.
\end{minipage}
\end{table*}

\section{Experiments}
\label{sec:experiments}

We evaluate the central premise of GRA: physical evidence should remain connected to interaction reasoning and executable behavior. The experiments examine overall driving performance, the contributions of the representation and staged learning, and how grounded cognition supports the generated actions.

\subsection{Setup}
\label{sec:setup}

GRAVA uses Qwen3-VL-8B-Instruct~\cite{bai2025qwen3vl} as its backbone and follows the GRA-aligned training procedure in Sec.~\ref{sec:active-rl}. Only about 60\% of the available human driving demonstrations are used for action supervision. Unless otherwise specified, all main results use a single greedy autoregressive completion per scene; multiple samples are used for self-distillation, diagnostic analyses, and Active RL. The generated Executable Planner action is converted to a trajectory by the fixed decoder. Benchmark definitions and evaluation protocols follow Sec.~\ref{sec:data_eval}.

\subsection{Main NAVSIM Results}
\label{sec:main-result}

We first assess the overall driving performance of the GRA framework. With a single camera and one greedy completion per scene, GRAVA achieves 90.48 PDMS on NAVSIM navtest. It obtains the highest PDMS among the directly autoregressive VLAs in Table~\ref{tab:navsim-context} and remains competitive with systems that introduce an additional trajectory head or world model. The table groups methods by model design to place this result in context, with differences in backbone, sensor inputs, and training data.

This performance is achieved by generating grounded reasoning and executable actions within one autoregressive policy, with no learned trajectory head or world model. Active RL further improves this policy using the driving outcomes of its generated sequences, while the visual input, greedy decoding procedure, and fixed trajectory decoder remain unchanged. We next examine how the representation and its training stages contribute to this performance.

\subsection{Component Ablation on NAVSIM}
\label{sec:navsim-ablation}

Table~\ref{tab:navsim-ablation} relates the overall performance to two aspects of the framework: learning grounded cognition and its action interface, and refining the resulting policy through post-training. All variants follow the same NAVSIM evaluation protocol.

GRA pre-training provides an important foundation for planning. Removing it reduces PDMS by 10.68 points, from 90.48 to 79.80. Figure~\ref{fig:training}(a) also shows lower loss at the start of planning training with GRA pre-training than when training the planner from a cold start. These observations support the role of grounded cognition before learning to use it for action generation. Replacing the Executable Planner with direct waypoint generation reduces PDMS by 3.25 points, indicating that the primitive-specific action representation helps translate the learned reasoning into vehicle motion.

Verified self-distillation provides a further 0.36-point improvement before RL. The resulting policy can already generate high-reward behaviors in some scenes, but does not yet produce them reliably. Active RL improves this policy from 82.10 to 90.48 PDMS by optimizing complete reasoning-to-action sequences using driving outcomes. Figure~\ref{fig:training}(b) shows gains across successive active loops, while continued training beyond an on-policy window without refreshing the active set leads to declining performance. This behavior supports refreshing the recoverable scenarios as the policy changes, as described in Sec.~\ref{sec:active-rl}. The internal benchmark further examines how this learning process affects decisions in complex interactions.

\begin{figure}[t]
\centering
\includegraphics[width=\linewidth]{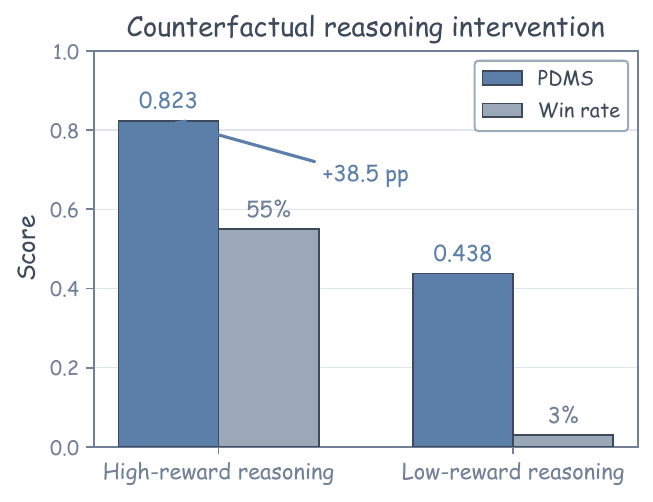}
\caption{Effect of reasoning intervention on action generation. Normalized PDMS and win rate are compared under high- and low-reward GRA reasoning, with the scene input and deterministic trajectory decoder held fixed.}
\label{fig:reasoning-intervention}
\end{figure}

\begin{table}[t]
\centering
\footnotesize
\setlength{\tabcolsep}{2.6pt}
\renewcommand{\arraystretch}{1.05}
\caption{Grounded question-answering performance on the held-out GR-NavSim evaluation set.}
\label{tab:qa-competence}

\begin{tabularx}{\linewidth}{@{}l *{4}{>{\centering\arraybackslash}X}@{}}
\toprule
Subtype & Qwen3-VL-8B & Kimi-K2.5 & GPT-5.4 & GRAVA \\
\midrule

\multicolumn{5}{@{}l}{\textit{Scene}} \\
\quad Summary & 6.09 & 6.03 & 7.08 & \textbf{7.47} \\
\quad Traffic sign & 4.74 & 5.58 & 5.87 & \textbf{7.63} \\
\quad Navigation & 6.21 & 6.44 & 6.33 & \textbf{6.67} \\
\quad Congestion & 5.41 & 6.79 & 6.59 & \textbf{7.92} \\
\quad \textit{Average} & 5.61 & 6.21 & 6.47 & \textbf{7.42} \\

\addlinespace
\multicolumn{5}{@{}l}{\textit{Object perception}} \\
\quad Critical object description & 2.01 & 2.19 & 2.79 & \textbf{5.71} \\
\quad Critical object identification & 1.32 & 2.80 & 3.71 & \textbf{5.69} \\
\quad \textit{Average} & 1.65 & 2.51 & 3.27 & \textbf{5.70} \\

\addlinespace
\multicolumn{5}{@{}l}{\textit{Object interaction}} \\
\quad Intention & 4.69 & 5.04 & \textbf{6.37} & 6.13 \\
\quad Navigation interaction & 4.20 & 4.60 & 6.17 & \textbf{6.49} \\
\quad Object interaction & 4.00 & 3.31 & 5.06 & \textbf{5.23} \\
\quad Map interaction & 4.68 & 4.65 & 5.62 & \textbf{6.73} \\
\quad Ego interaction & 4.27 & 4.03 & 4.94 & \textbf{6.00} \\
\quad \textit{Average} & 4.37 & 4.32 & 5.63 & \textbf{6.12} \\

\addlinespace
\multicolumn{5}{@{}l}{\textit{Decision (label accuracy)}} \\
\quad Object decision & 3.16 & 3.68 & 4.21 & \textbf{8.34} \\
\quad Ego decision & 6.05 & 6.32 & 5.13 & \textbf{8.66} \\

\midrule
\textbf{Overall} & 4.40 & 4.77 & 5.39 & \textbf{6.86} \\
\bottomrule
\end{tabularx}
\par\smallskip
\begin{minipage}{\linewidth}
\footnotesize\raggedright
Scene, object-perception, and object-interaction subtypes use $0$--$10$ scores; decision subtypes use label accuracy. Average and Overall denote category and aggregate scores, respectively. Qwen3-VL-8B is the untuned backbone. Bold indicates the best result; higher is better.
\end{minipage}
\end{table}

\subsection{Internal Long-Tail Results}
\label{sec:internal-result}

The internal benchmark evaluates decision-making and the ability to navigate complex long-tail scenes involving route obstructions, lane borrowing, and road hazards. The ego vehicle must decide when to proceed or wait and plan a safe, feasible trajectory that carries out that decision. Following Sec.~\ref{sec:internal-benchmark}, Key-Object Compliance (KOC) measures the execution of interaction decisions, while CDS evaluates overall driving performance. We vary the pre-training target, reasoning structure, and use of Active RL while keeping inference fixed, with results reported in Table~\ref{tab:internal-result}.

Grounded pre-training provides stronger benefits than ungrounded cognition supervision. Ungrounded pre-training improves CDS by 2.5 points over no pre-training, while GRA pre-training further improves CDS by 7.9 points and KOC by 8.1 points. This comparison extends the NAVSIM ablation by showing the value of grounding the cognition learned before planning, particularly for responding to interaction-critical objects.

The reasoning structure provides benefits beyond object grounding alone. Full GRA improves KOC and CDS by 14.9 and 15.3 points over action-only prediction, and by 4.4 and 4.6 points over the grounded-objects-only variant. The latter comparison directly addresses the role of the interaction and decision paths in Sec.~\ref{sec:gra}: identifying an object supplies the evidence, while reasoning about its effect on the ego route supports decisions about whether and how to proceed. The gains in both metrics indicate better execution of these decisions and improved driving performance through the scene.

Active RL further improves the policy's ability to carry out these decisions in complex long-tail scenes. KOC increases from 69.8 to 92.3 and CDS from 73.1 to 90.1, whereas Progress and Comfort improve by 2.8 and 0.7 points. These gains reflect better execution of appropriate pass-or-hold behavior and improved overall driving performance. The larger increase in KOC indicates that the benefit mainly lies in carrying out the appropriate maneuver, rather than simply advancing farther through the scene.

\subsection{Grounded Reasoning Analysis}
\label{sec:grounding-analysis}

The long-tail results motivate a closer examination of how GRA reasoning uses physical evidence and influences the action generated from it. We also evaluate the grounded cognition capabilities that support this process.

\begin{figure*}[!t]
\centering
\includegraphics[width=\textwidth]{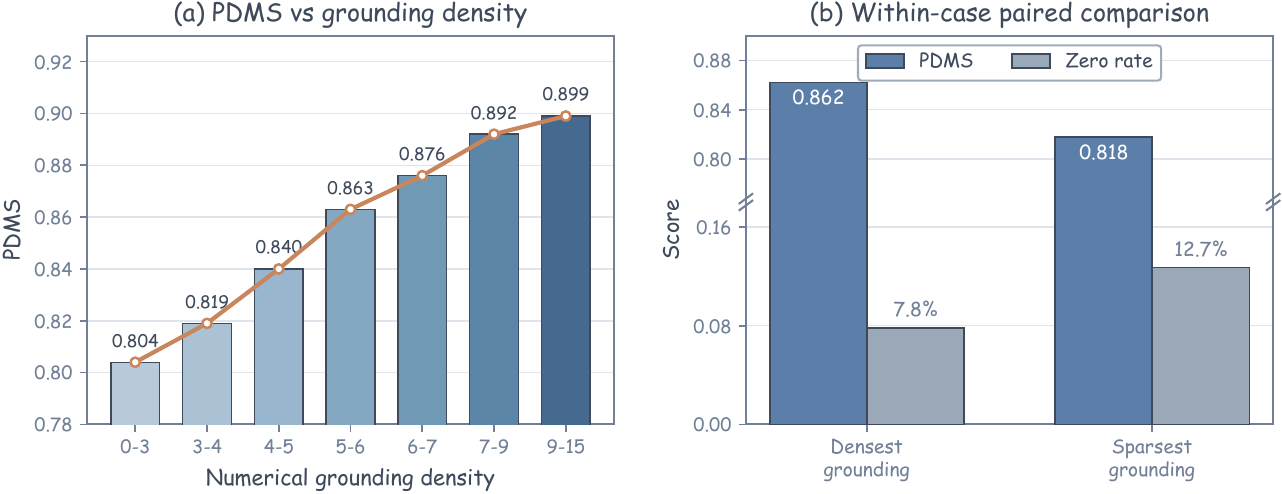}
\caption{Numerical grounding density and planning quality. Density counts references to action-relevant object distances and speeds. (a) Mean normalized PDMS across density bins. (b) Within-scene comparison of the densest and sparsest completions in normalized PDMS and zero-PDMS rate.}
\label{fig:numerical-grounding}

\par\bigskip
\centering
\includegraphics[width=\textwidth]{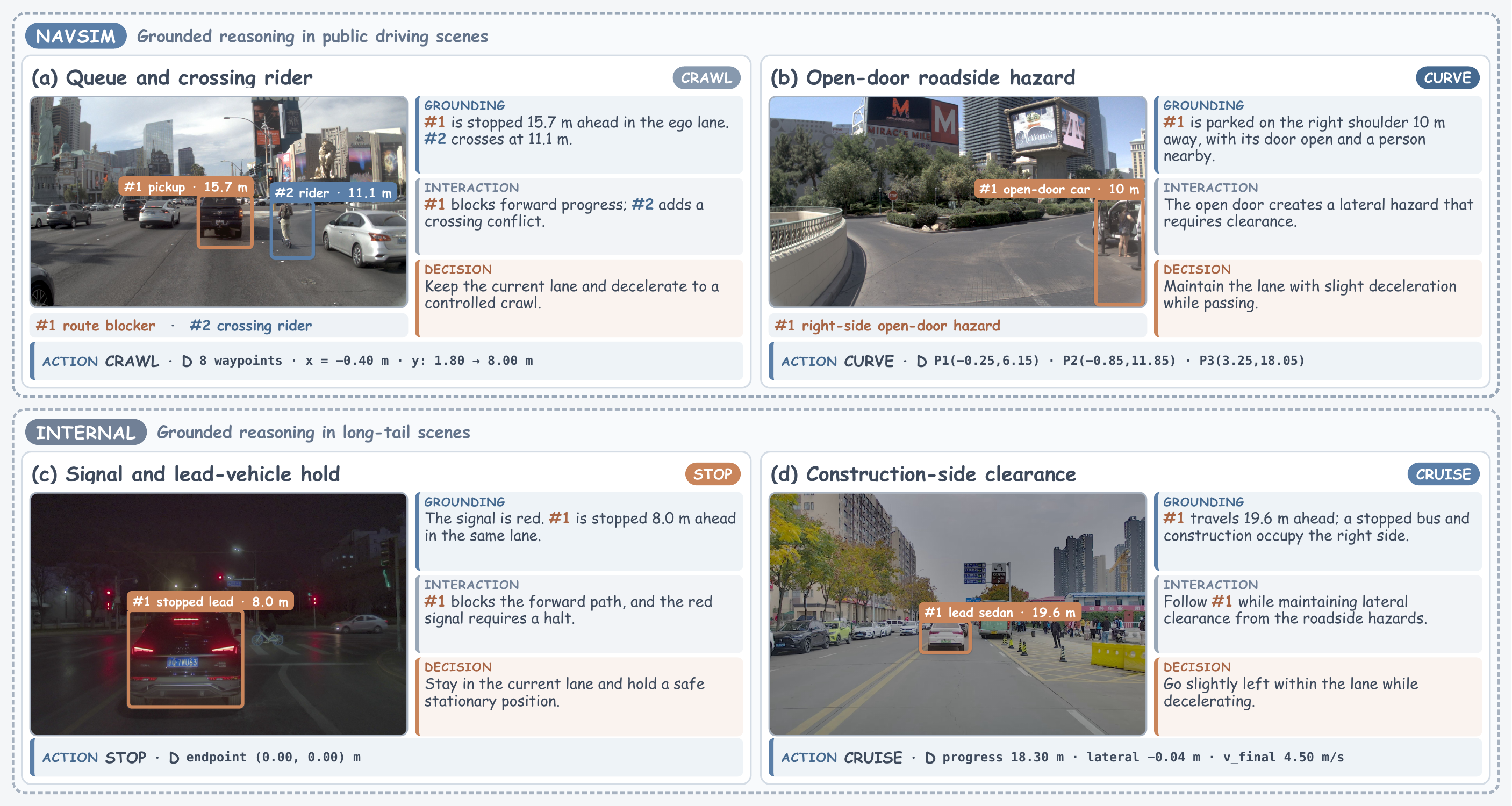}
\caption{Qualitative GRA examples on NAVSIM (top) and the internal long-tail benchmark (bottom). Each example links grounded scene evidence to interaction reasoning, driving decisions, and an Executable Planner action.}
\label{fig:qualitative}
\end{figure*}

\paragraph{Effect of Reasoning on Action Generation}
GRA learns action generation as a continuation of grounded reasoning (Sec.~\ref{sec:overview}). To examine whether this preceding reasoning affects behavior, we hold the scene input and deterministic trajectory decoder fixed and replace only the reasoning sequence. As shown in Fig.~\ref{fig:reasoning-intervention}, replacing low-reward reasoning with its high-reward counterpart changes the predicted action and increases normalized PDMS from 0.438 to 0.823, while the win rate rises from 3\% to 55\%. This intervention provides direct evidence that the generated action depends on the reasoning in its context, supporting the reasoning-to-action connection proposed to address \textbf{G2}.

\paragraph{Numerical Grounding}
Having established that reasoning affects action generation, we examine the physical information used within it. Numerical grounding density counts references to the distances and speeds of action-relevant objects, corresponding to the physical-state evidence introduced to address \textbf{G1}. Figure~\ref{fig:numerical-grounding}(a) shows that normalized PDMS increases from 0.804 in the lowest-density bin to 0.899 in the highest. The within-scene comparison in Fig.~\ref{fig:numerical-grounding}(b) follows the same trend: the densest completion averages 0.862 normalized PDMS with a 7.8\% zero-score rate, compared with 0.818 and 12.7\% for the sparsest. Both comparisons associate explicit physical-state grounding with better planning outcomes. The measured content concerns object distances and speeds, rather than the overall length of the reasoning trace.

\paragraph{Grounded Question-Answering Performance}
The preceding analyses examine reasoning during planning; the held-out GR-NavSim QA set evaluates the individual cognition capabilities underlying that reasoning. This follows the shared-supervision design in Sec.~\ref{sec:gra}, where questions teach grounded concepts that reappear in the complete reasoning-to-action sequence. We compare GRAVA with GPT-5.4~\cite{openai2026gpt54thinking}, Kimi-K2.5~\cite{kimiteam2026kimi25}, and the untuned Qwen3-VL-8B-Instruct backbone. Open-ended responses are scored by an independent LLM judge, while decision questions are evaluated using label accuracy. All models receive the same ego-frame coordinate prompt for localization.

GRAVA achieves the highest overall score of 6.86 and leads on 12 of the 13 subtypes (Table~\ref{tab:qa-competence}); GPT-5.4 performs slightly better only on intention prediction. The strongest gains over the untuned backbone concern critical-object perception and object-level decisions. For example, the critical-object identification score rises from 1.32 to 5.69, while the object-decision score rises from 3.16 to 8.34. These improvements concern recognizing the relevant physical evidence and determining the response required by each object. Together with the planning ablations, they support learning these grounded concepts through shared supervision and using them in subsequent action generation.

\subsection{Executable Planner Analysis}
\label{sec:planner-analysis}

Grounded cognition must ultimately be expressed as vehicle motion. The Executable Planner ablation establishes the benefit of this action interface; we now examine how its parameters account for differences in trajectory quality. Among low-score cases with a large reward gap between sampled actions, high- and low-reward trajectories use the same motion primitive in most cases. Their differences lie in endpoints, lateral offsets, speed profiles, and \texttt{CRAWL} waypoints. Selecting the motion regime therefore leaves an important part of the driving decision to be resolved through its continuous parameters.

Within the same motion primitive, driving rewards distinguish parameter choices that lead to different trajectory quality. The primitive-specific action space in Sec.~\ref{sec:planner} therefore allows Active RL to refine how a behavior is executed through its endpoints, lateral offsets, or speed parameters. This complements the improved compliance with interaction constraints observed on the internal benchmark.

\subsection{Qualitative Grounded Reasoning-to-Action Analysis}

Figure~\ref{fig:qualitative} brings these analyses together through complete GRA examples. In the NAVSIM queue scene, grounding identifies a stopped vehicle in the ego lane and a crossing rider. The interaction reasoning assigns different constraints to these objects: the vehicle blocks forward progress, while the rider introduces a crossing conflict. Their combined effect supports keeping the lane and decelerating to a controlled crawl, which is expressed by the CRAWL action. The same object evidence is thus preserved from reference resolution through interaction reasoning to executable behavior.

The remaining cases illustrate different consequences of this structure. An open-door roadside hazard requires lateral clearance and produces a CURVE trajectory. On the internal benchmark, a red signal and stopped lead vehicle require a STOP action, whereas a moving lead vehicle and roadside construction call for following with lateral clearance, expressed through CRUISE parameters. Across these cases, GRA connects what the model observes to why a response is needed and how it is executed.


\section{Conclusion and Future Work}
\label{sec:conclusion}

We presented GRAVA, a Grounded Reasoning-to-Action framework that connects physical scene evidence, structured reasoning, and executable behavior through a shared GRA representation. GRA binds action-relevant linguistic references to visual regions and physical states, while organizing interactions, decisions, and planning anchors in a trajectory-anchored graph. The same representation supports both cognition and planning supervision, while GRA-aligned pre-training, the Executable Planner, verified self-distillation, and Active RL progressively transform grounded reasoning into executable actions. The resulting GRA-centered data pipeline further produces GR-NavSim with 107K grounded scenes and 2.2M grounded question-answer pairs. GRAVA-8B achieves 90.48 PDMS on NAVSIM, while full GRA improves Key-Object Compliance and Closed-loop Driving Score by 19.3\% and 20.5\% relative to action-only prediction, respectively. These results demonstrate the benefit of constructing grounding, reasoning, and action learning around a shared action-relevant representation.

Future work will extend GRAVA toward temporal and multi-view grounding, interactive closed-loop environments, longer-horizon behaviors, and more efficient reasoning through distillation and adaptive computation.


\bibliographystyle{IEEEtran}
\bibliography{reference,bibliography/ieee_style_control}

\newpage

\begin{IEEEbiography}
[{\includegraphics[width=1in,height=1.25in,clip,keepaspectratio]{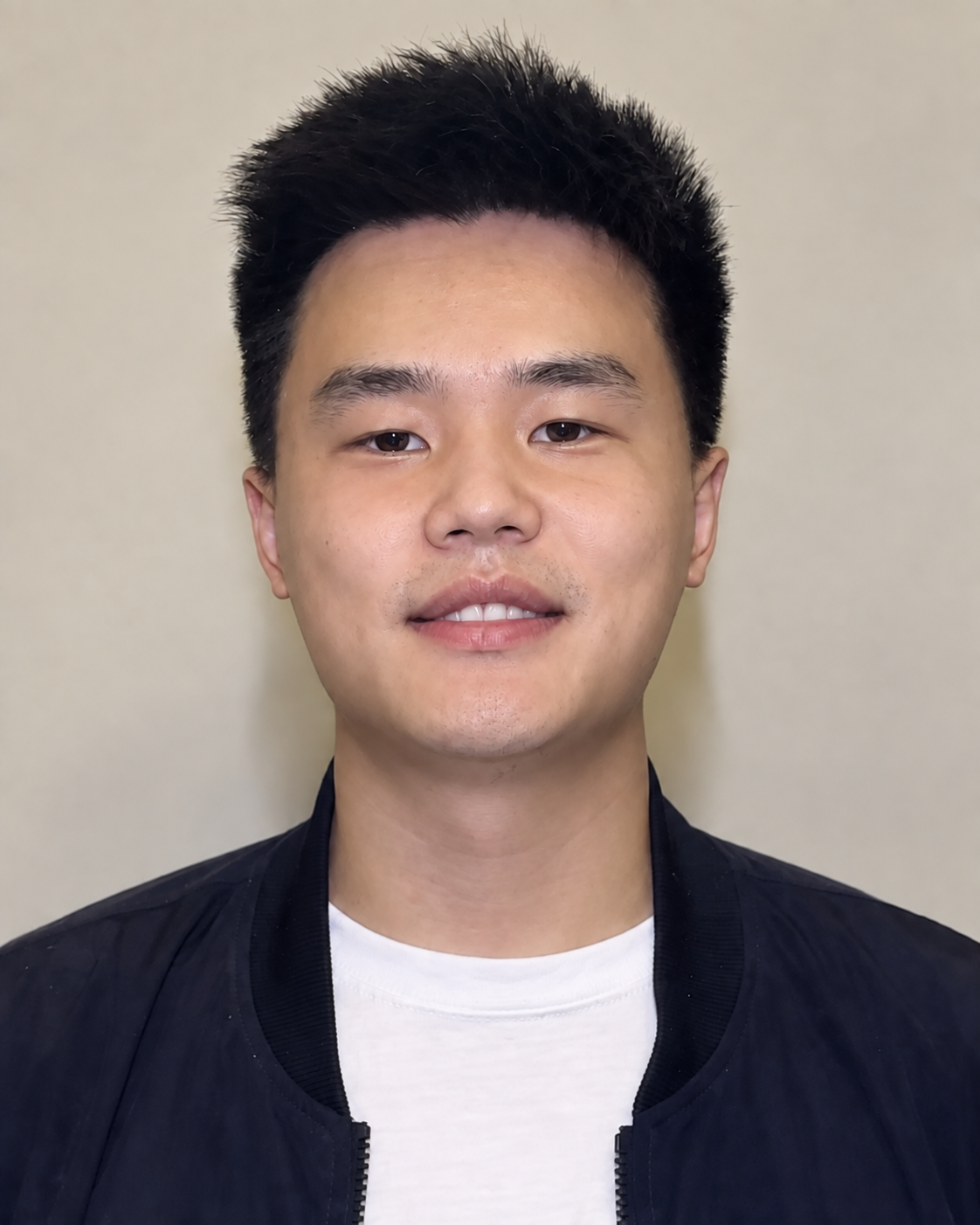}}]
{Xiao Liu} received the M.S. degree in vehicle engineering from the Harbin Institute of Technology, Harbin, China, in 2021. He is pursuing the Ph.D. degree in intelligent vehicle technology at the Beijing Institute of Technology, Beijing, China. He was a research intern at Baidu and Shineon AI, working on large vision-language models for autonomous driving. His research interests include vision-based perception, end-to-end autonomous driving, and large language models for complex driving scenarios.
\end{IEEEbiography}

\begin{IEEEbiography}
[{\includegraphics[width=1in,height=1.25in,clip,keepaspectratio]{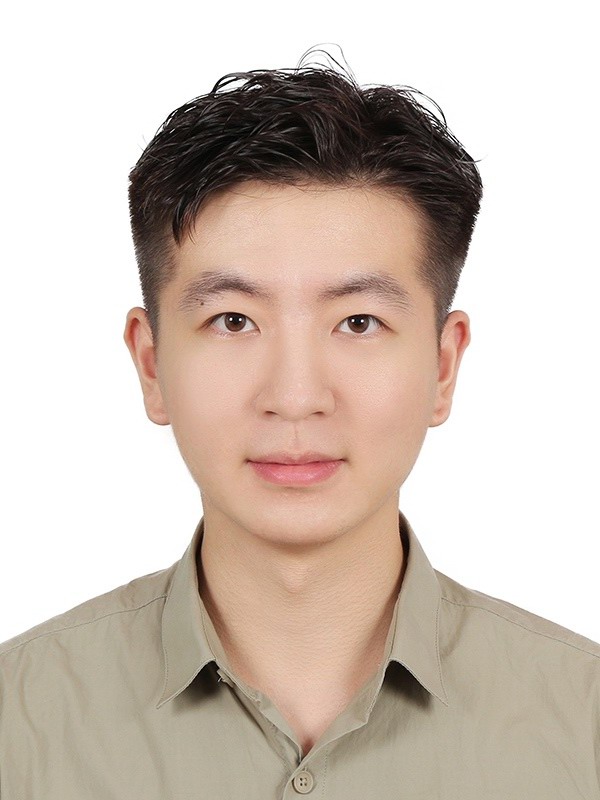}}]
{Haoyu Li} is currently pursuing the Ph.D. degree at the National Engineering Research Center of Electric Vehicles, Beijing Institute of Technology, China. He is also a visiting Ph.D. student at the School of Electrical and Electronic Engineering, Nanyang Technological University, Singapore. His research interests include robust multi-sensor fusion for perception and prediction in autonomous driving, and embodied intelligence for physical-world agents.
\end{IEEEbiography}

\begin{IEEEbiography}
[{\includegraphics[width=1in,height=1.25in,clip,keepaspectratio]{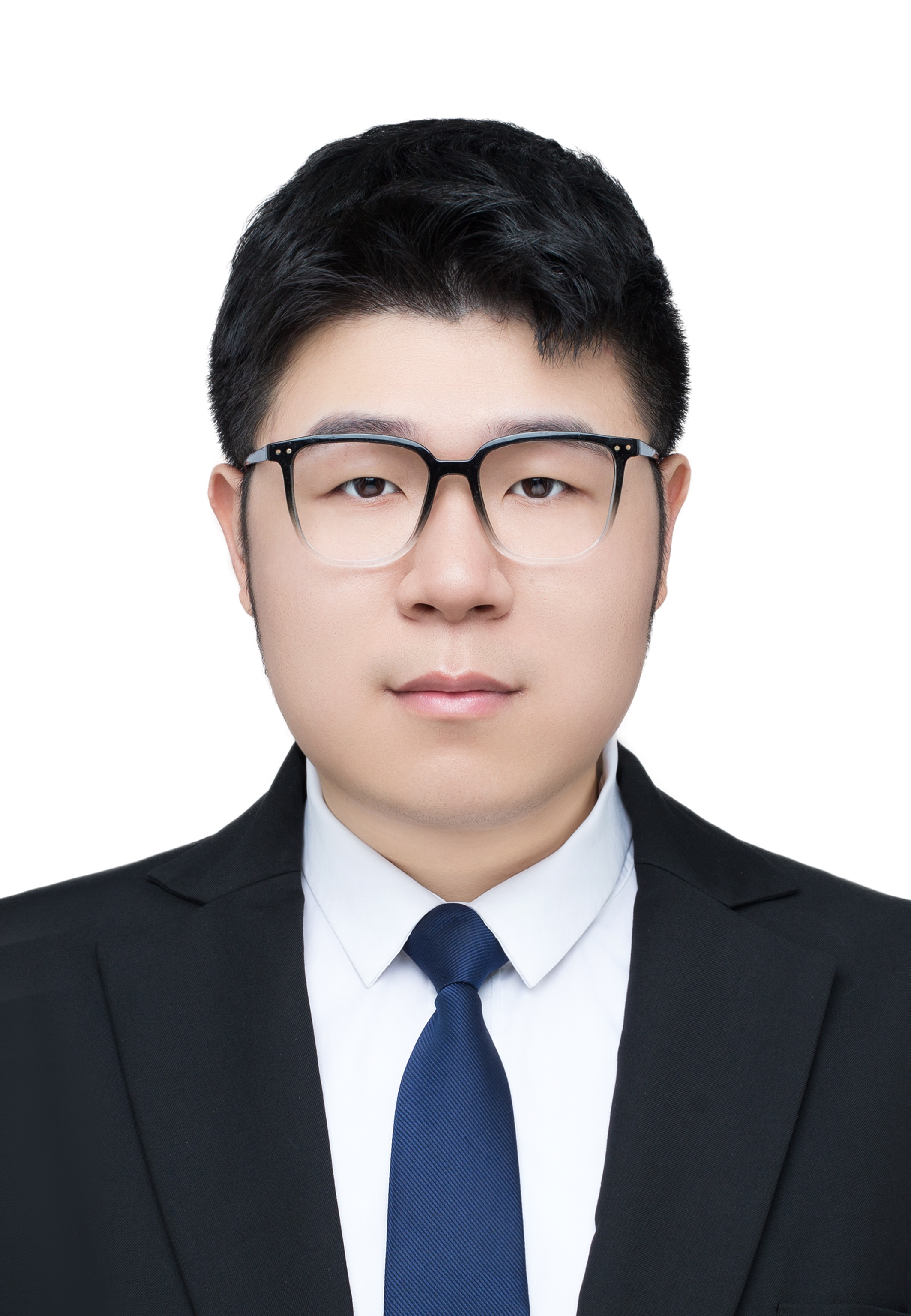}}]
{Jianghao Leng} received the B.S. degree in vehicle engineering from Beijing Institute of Technology, Beijing, China, in 2019, and the Ph.D. degree in mechanical engineering from Beijing Institute of Technology in 2025. From 2023 to 2024, he was a visiting student at the College of Design and Engineering, National University of Singapore. His research interests include multi-sensor fusion SLAM, motion planning and eco-driving for connected and autonomous vehicles (CAVs).
\end{IEEEbiography}

\begin{IEEEbiography}
[{\includegraphics[width=1in,height=1.25in,clip,keepaspectratio]{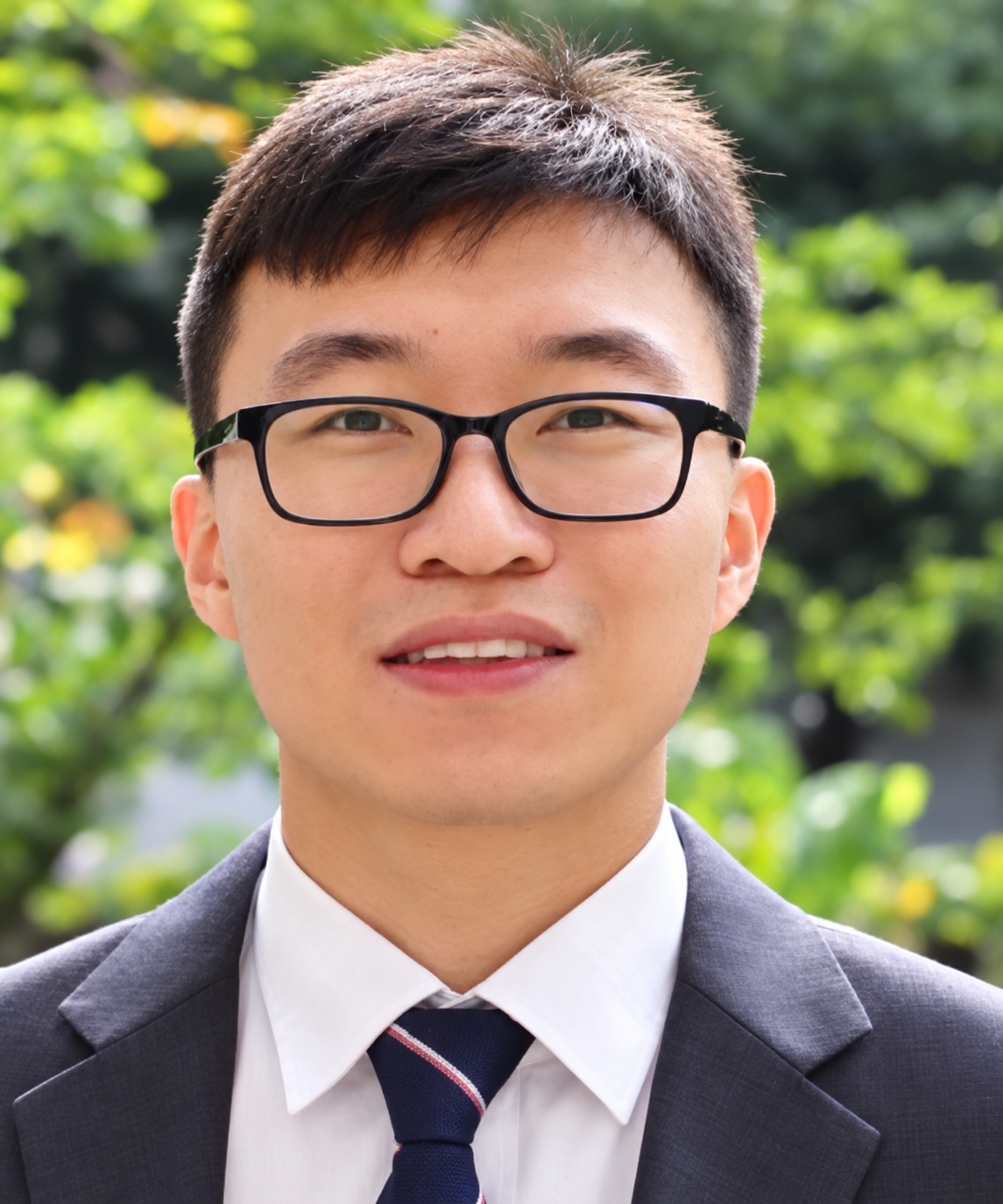}}]
{Lin Wang} (Member, IEEE) is currently an Assistant Professor with the School of Electrical and Electronic Engineering, Nanyang Technological University, Singapore, where he leads the Embodied Perception and InterACTion (EmPACT) Lab. His research interests include bio-inspired sensing and perception, event-based vision, sensor fusion, neuromorphic multimodal AI, spatial intelligence, embodied robotic systems, and edge AI. 
\end{IEEEbiography}

\begin{IEEEbiography}
[{\includegraphics[width=1in,height=1.25in,clip,keepaspectratio]{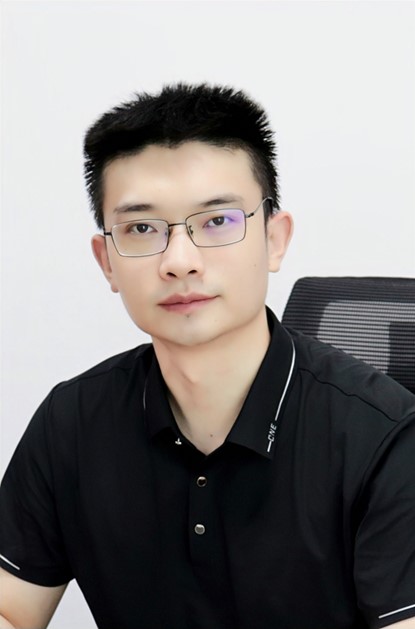}}]
{Chao Sun} (Member, IEEE) received the B.S. and Ph.D. degrees in mechanical engineering from Beihang University, Beijing, China, and the Beijing Institute of Technology, Beijing, China, in 2010 and 2016, respectively. He was a postdoctoral researcher with the Energy, Controls, and Applications Laboratory, University of California, Berkeley, CA, USA. He is currently an Associate Professor at the Beijing Institute of Technology. His research interests include automated and connected vehicles and hybrid electric vehicles.
\end{IEEEbiography}

\clearpage
\begingroup
\appendices
\setcounter{figure}{0}
\setcounter{table}{0}
\setcounter{algorithm}{0}
\setcounter{equation}{0}
\setcounter{topnumber}{2}
\renewcommand{\thefigure}{S\arabic{figure}}
\renewcommand{\thetable}{S\arabic{table}}
\renewcommand{\thealgorithm}{S\arabic{algorithm}}
\renewcommand{\theequation}{S\arabic{equation}}
\renewcommand{\theHfigure}{supplement.\arabic{figure}}
\renewcommand{\theHtable}{supplement.\arabic{table}}
\providecommand{\theHalgorithm}{}
\renewcommand{\theHalgorithm}{supplement.\arabic{algorithm}}
\renewcommand{\theHequation}{supplement.\arabic{equation}}

\renewcommand{\ttdefault}{ptm}
\renewcommand{\encodingdefault}{T1}
\fontencoding{T1}\selectfont

\graphicspath{{appendix/figures/}}
\interdisplaylinepenalty=2500
\hyphenation{op-tical net-works semi-conduc-tor IEEE-Xplore}

\lstset{
basicstyle=\rmfamily\scriptsize, columns=fullflexible, keepspaces=true, breaklines=true, frame=single, xleftmargin=2pt, xrightmargin=2pt }

\markboth{SUPPLEMENTARY MATERIAL, IEEE TRANSACTIONS ON PATTERN ANALYSIS AND MACHINE INTELLIGENCE}%
{Liu \MakeLowercase{\textit{et al.}}: Supplementary Material}

\section{Implementation Details}
\label{app:reproducibility}

All stages of the progressive GRA training procedure in Sec.~\ref{sec:active-rl} fine-tune Qwen3-VL-8B-Instruct with DeepSpeed ZeRO-2. Each input contains the native $1920\times1080$ front-camera image, ego state, motion history, and navigation command. Table~\ref{tab:train-config} gives the training settings.

\begin{table}[H]
\centering
\scriptsize
\setlength{\tabcolsep}{2pt}
\caption{Training configuration. Batch size is per GPU. GA denotes gradient accumulation.}
\label{tab:train-config}
\begin{adjustbox}{max width=\columnwidth}
\begin{tabular}{@{}l c c c c l@{}}
\toprule
Stage & Epochs & LR & Batch / GA & Max length & Compute \\
\midrule
GRA pre-training & 1 & $1{\times}10^{-5}$ & 4 / 1 & 4,096 & $4{\times}8$ H20 \\
Planner warm-up & 3 & $1{\times}10^{-5}$ & 4 / 1 & 4,096 & $2{\times}8$ H20 \\
Self-distillation & 1 & $5{\times}10^{-6}$ & 4 / 1 & 4,096 & $2{\times}8$ H20 \\
Active RL & \textemdash & $1{\times}10^{-6}$ & 4 / 1 & 4,096 & \shortstack[l]{1 rollout + 3 training\\+ 1 sim node} \\
\bottomrule
\end{tabular}
\end{adjustbox}
\end{table}

\paragraph*{NAVSIM data composition} Of the 107K grounded GR-NavSim scenes reported in the main paper, 102,861 have valid official evaluation caches and form the planner and Active RL pool. Grounding-consistency filtering retains 62,111 of these cases for planner warm-up.

\subsection{Verified Self-Distillation}
\label{app:self-distill}

Starting from the planner warm-up policy, GRAVA samples complete reasoning-to-action sequences $y_i=[r_i;a_i]$ and selects candidates using the closed-loop rewards of their decoded trajectories. As defined in Eq.~\eqref{eq:sd-target}, each retained target $y_i^{+}=[r_i^{+};a_i^{+}]$ keeps the reasoning and Executable Planner action from the same reward-selected rollout. The distilled policy initializes Active RL and serves as its frozen reference.

\subsection{Executable Planner Decoding}
\label{app:planner-decoder}

Algorithm~\ref{alg:planner-decoder} defines the fixed decoder $D_p(g,\phi;s_{\mathrm{ego}})$ that realizes each autoregressive Executable Planner action as an eight-waypoint ego-frame trajectory. Each branch reads the fields declared in Table~\ref{tab:planner_schema}.

\begin{algorithm}[!t]
  \caption{Executable Planner decoding
  $\hat{\tau}=D_p(g,\phi;s_{\mathrm{ego}})$.}
  \label{alg:planner-decoder}
\begin{lstlisting}[frame=none]
Input : structured action
        {planner: p, gear: g, params: phi},
        ego state s_ego
Output: eight-waypoint trajectory
        tau_hat = (x_1..8, y_1..8)

 1: parse p, g, phi <- action
 2: assert phi in Phi_p              # validate schema
 3: p0 <- (0, 0)                     # ego origin
 4: switch p do
 5:   case STOP:
 6:     q <- stop_profile(||phi.endpoint||,
 7:                       speed(s_ego), 8)
 8:     tau_hat <- q * phi.endpoint
 9:   case CRAWL:
10:     tau_hat <- decode_points(g, phi.points)
11:   case CURVE:
12:     B(t) <- Bezier(
13:       p0, decode_controls(g, phi))
14:     tau_hat <- {B(k / 8)} for k = 1..8
15:   case CRUISE:
16:     y <- monotone_Hermite(
17:       phi.progress_end, speed(s_ego),
18:       phi.v_end, 8)
19:     x <- {k * phi.lat_end / 8} for k = 1..8
20:     tau_hat <- {(x_k, y_k)} for k = 1..8
21: return tau_hat                    # ego frame
\end{lstlisting}
\end{algorithm}

The decoder has no learned parameters. \texttt{STOP} expands an endpoint into a smooth profile based on the current ego speed, and \texttt{CRAWL} decodes eight position controls. \texttt{CURVE} evaluates a cubic B\'ezier curve from the ego origin $p_0=(0,0)$,
\begin{equation}
  B(t)=\sum_{j=0}^{3}\binom{3}{j}(1-t)^{3-j}t^j p_j,
  \qquad t\in[0,1],
  \label{eq:bezier}
\end{equation}
at $t\in\{1/8,\ldots,1\}$. \texttt{CRUISE} uses a cubic Hermite profile for monotone longitudinal progress and linear interpolation for the terminal lateral offset. The current and terminal speeds condition the longitudinal profile.

\paragraph*{Why \texttt{CRAWL} emits explicit waypoints} \texttt{CRAWL} outputs eight points for low-speed stop-and-go motion. Its coordinate range is narrower than that of unrestricted waypoint regression, so direct prediction remains tractable. The points retain small longitudinal changes that a low-dimensional curve may smooth out. The other primitives span wider motion ranges and use compact geometric parameters.

\subsection{Active RL Configuration}
\label{sec:active-analysis}

Active RL scores each complete GRA reasoning-to-action sequence with the PDMS of its decoded trajectory. We sample $K=8$ completions at temperature $0.8$ and top-$p=0.95$, set the KL coefficient to $\beta=0.01$, and use DAPO-style asymmetric clipping at $(0.20,0.28)$. Rewards are normalized within each group; overlong completions are filtered, with up to three resampling attempts.

\paragraph*{Recoverable scenario selection} We screen every training case with one greedy completion, yielding a mean normalized PDMS of 0.9101. Cases below 0.9 receive eight sampled rollouts. We retain cases that satisfy all three reward-variation criteria:
\begin{equation}
  \min s_i < 0.8,\qquad
  \max s_i\ge0.9,\qquad
  \operatorname{std}(s_i)\ge0.10
  \label{eq:active-criteria}
\end{equation}
Table~\ref{tab:active-set} reports the number of cases after each filter.

\begin{table}[H]
  \centering
  \small
  \caption{Selection of recoverable scenarios for the Active RL update set.}
  \label{tab:active-set}
  \begin{tabularx}{\linewidth}{@{}>{\raggedright\arraybackslash}X r >{\raggedright\arraybackslash}X@{}}
    \toprule
    Selection step & Cases & Meaning \\
    \midrule
    Current checkpoint evaluation & 102,861 & One greedy completion per case \\
    Greedy below 0.9 & 26,151 & Cases with potential for improvement \\
    Best-of-8 reaches at least 0.9 & 14,398 & Recoverable under best-of-8 sampling \\
    All reward-variation criteria & 4,676 & Cases used for the RL update \\
    \bottomrule
  \end{tabularx}
\end{table}

Low greedy reward alone does not define the training set. We also require low- and high-reward sampled outcomes and sufficient variance. Consecutive active sets of 4,676 and 6,464 cases share only 1,995 cases, a Jaccard overlap of 21.8\%, motivating rescreening after every outer loop.

\onecolumn
\section{Visualization of GRA Planning}
\label{app:gra-visualization}

\begin{figure}[H]
  \centering
  \includegraphics[width=\textwidth,height=0.8\textheight,keepaspectratio]{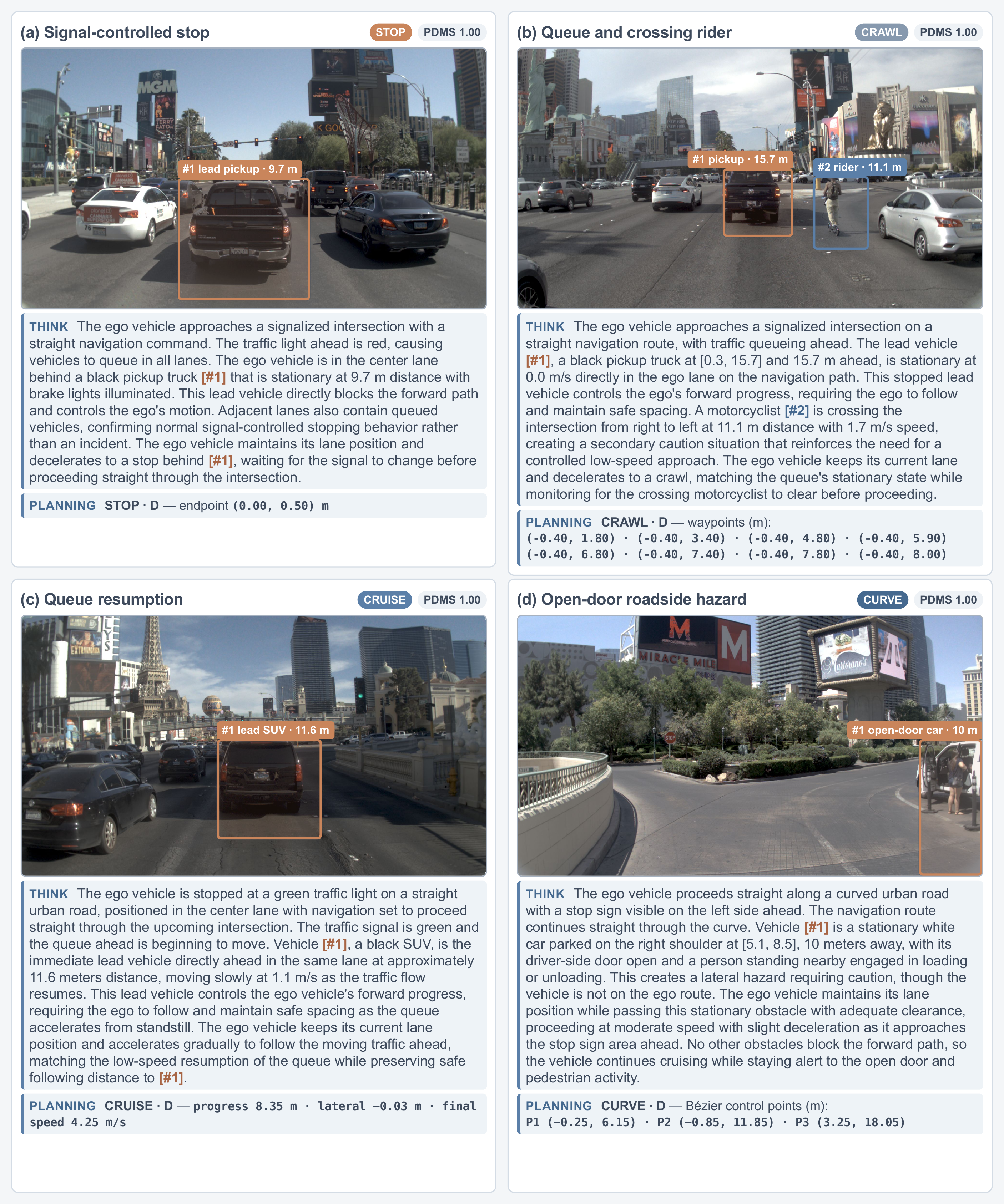}
  \caption{Expanded qualitative GRA planning examples for Fig.~\ref{fig:qualitative}
  on NAVSIM: (a) signal-controlled
  \texttt{STOP}, (b) \texttt{CRAWL} for a queue and crossing rider, (c)
  \texttt{CRUISE} as the queue resumes, and (d) \texttt{CURVE} around an open
  roadside door.}
  \label{fig:gra-navsim}
\end{figure}

\clearpage
\begin{figure}[H]
  \centering
  \includegraphics[width=\textwidth,height=0.8\textheight,keepaspectratio]{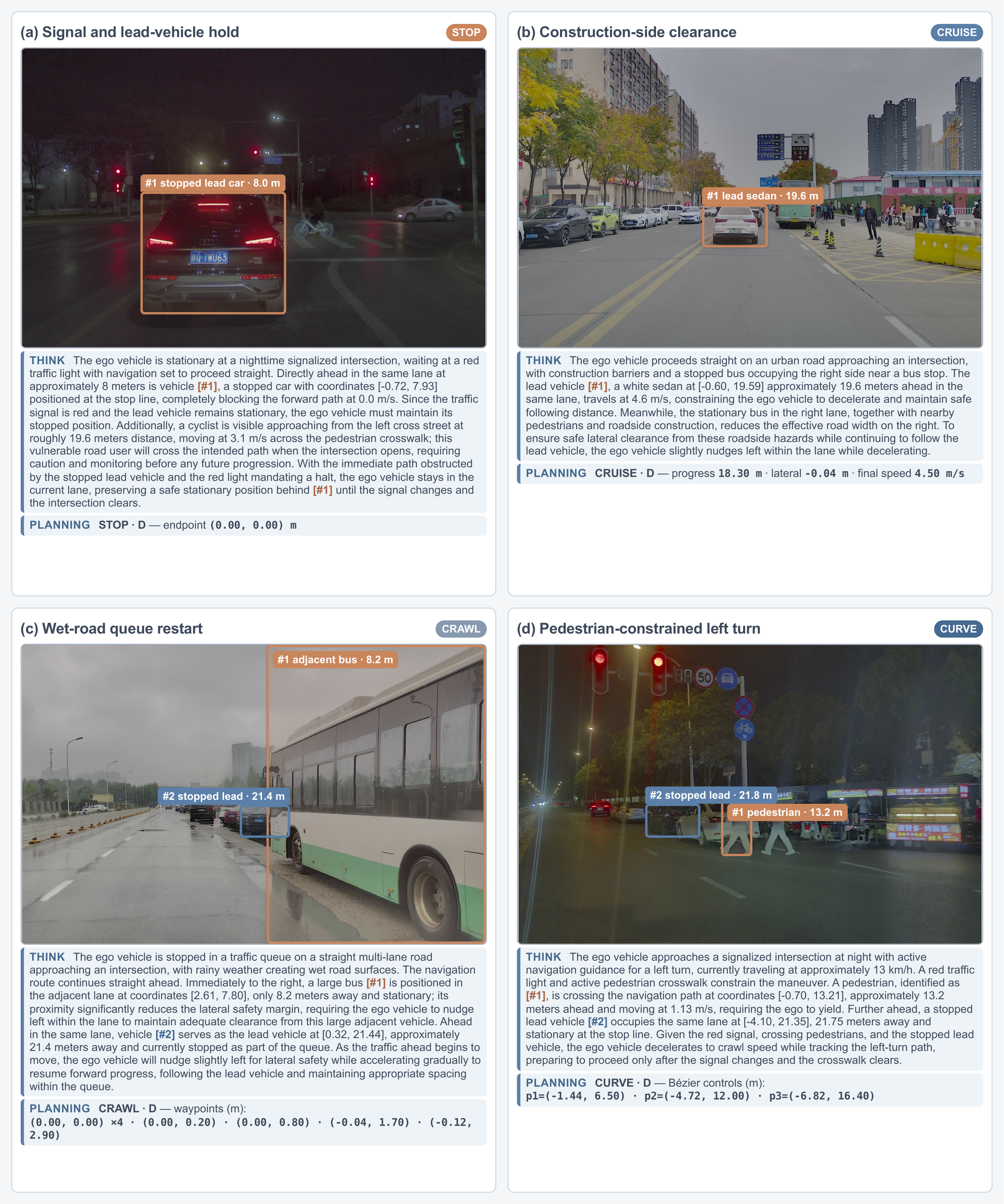}
  \caption{Qualitative GRA planning on the internal benchmark: (a)
  \texttt{STOP} behind a lead vehicle at a red signal, (b) \texttt{CRUISE}
  with lateral clearance from a construction zone, (c) \texttt{CRAWL} in a
  wet-road queue, and (d) \texttt{CURVE} for a left turn constrained by a
  pedestrian.}
  \label{fig:gra-internal}
\end{figure}

\twocolumn
\raggedbottom
\begingroup
\linespread{0.93}\selectfont
\setlength{\intextsep}{6pt}
\setlength{\textfloatsep}{8pt}
\setlength{\abovedisplayskip}{6pt}
\setlength{\belowdisplayskip}{6pt}
\setlength{\abovedisplayshortskip}{3pt}
\setlength{\belowdisplayshortskip}{3pt}
\section{Active RL Infrastructure}
\label{app:rl-infra}

Active RL requires trajectory-level rewards for complete autoregressive sequences. We therefore separate rollout generation, simulator scoring, and policy optimization into three services while keeping the reasoning and action tokens within the same rollout.

\begin{figure}[H]
  \centering
  \includegraphics[width=\linewidth]{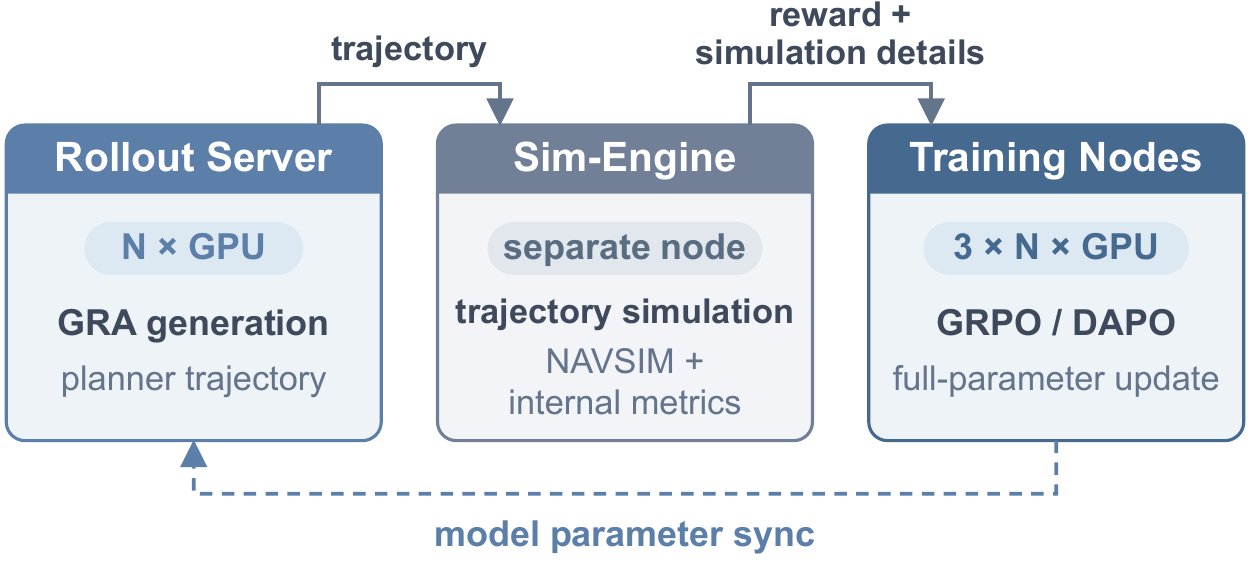}
  \caption{Distributed infrastructure for Active RL. The rollout server sends
  decoded trajectories to the sim-engine, which returns rewards and simulation
  metadata to the training nodes. After each update, the training nodes
  synchronize the model with the rollout server.}
  \label{fig:rl-infra}
\end{figure}

The rollout server runs vLLM on one node with eight H20 GPUs, tensor parallelism, a 4,096-token limit, and prefix caching. The trainer updates all parameters on three nodes with eight H20 GPUs each using PyTorch distributed data parallelism and DeepSpeed ZeRO-2. A separate sim-engine node stores scene caches and runs the dataset-specific evaluators.

\paragraph{Execution flow} For each prompt, the rollout server samples eight reasoning-to-action sequences with the settings in Section~\ref{sec:active-analysis}. The fixed decoder converts each valid action into an eight-waypoint trajectory. The server sends the trajectory and the scene metadata required by the evaluator to the sim-engine.

The sim-engine returns a scalar reward and simulation details keyed by the request identifier. The trainer updates a prompt group after the sim-engine scores all eight completions. An unparsable action receives zero reward. The trainer then normalizes rewards within the group and applies Eq.~\eqref{eq:rl-objective} of the main paper. It filters overlong completions and resamples zero-variance groups under the limits in Section~\ref{sec:active-analysis}. The training nodes synchronize the model with the rollout server after each update.

\paragraph{Simulator scoring API} The sim-engine uses one request format for two scoring modes. The NAVSIM and internal modes return the aggregate and component scores defined in Sections~\ref{sec:navsim-protocol} and~\ref{sec:internal-benchmark} of the main paper, respectively. The two modes share the rollout and training code but use different scene loaders and score adapters.

The reward client caches each simulator response so all reward components reuse one evaluation. Active-set screening, online RL, and checkpoint evaluation use the same endpoints. The sim-engine runs independently of distributed training.

\section{Annotation-Time Aliases and Training-Time Grounding}
\label{app:object-referencing}

\begin{figure}[!t]
  \centering
  \includegraphics[width=\linewidth]{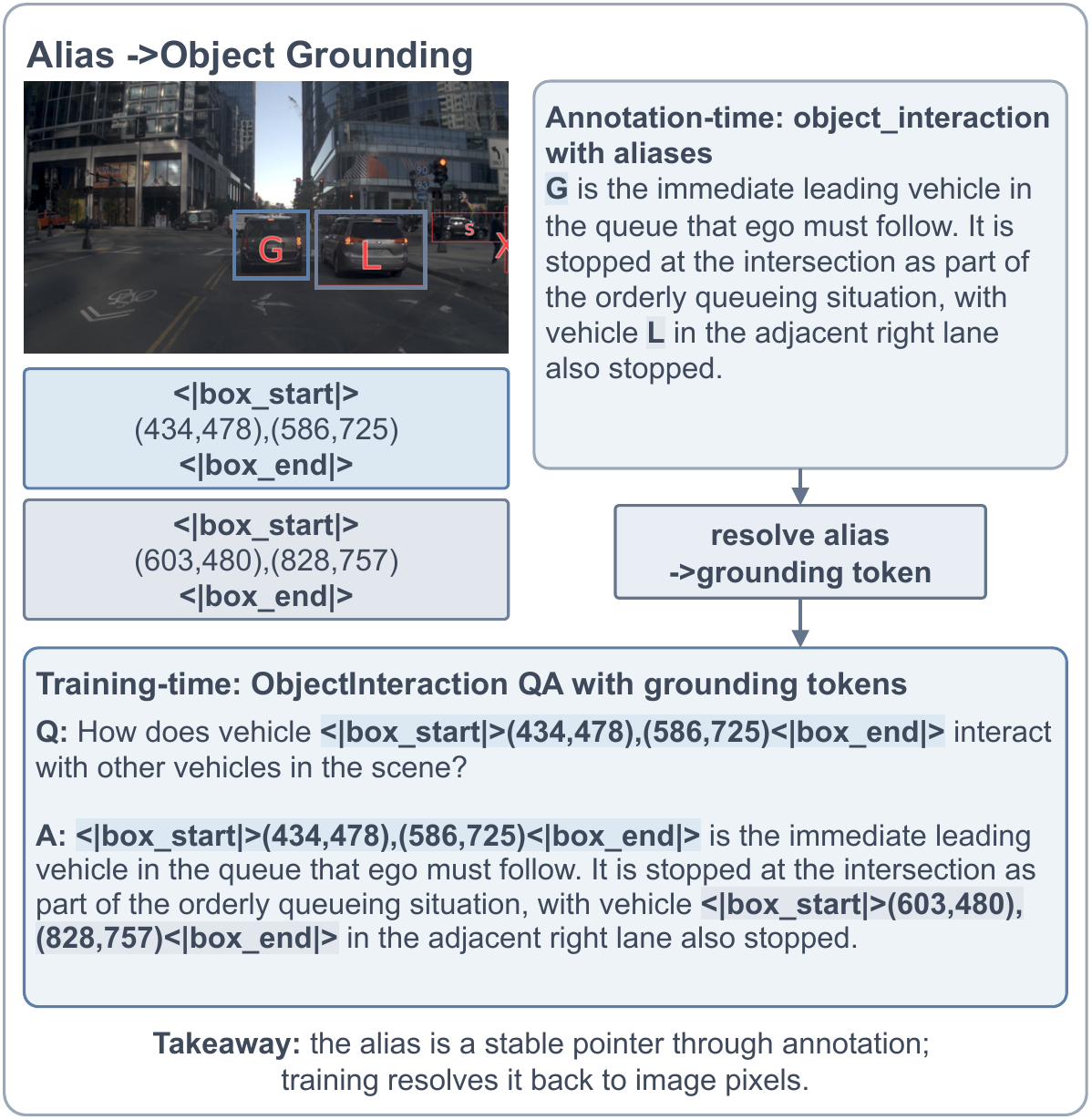}
  \caption{Alias resolution from annotation to training. Short aliases maintain
  object identity across annotation steps; training replaces them with
  bounding-box tokens referring to the same image regions, preserving the
  interaction content.}
  \label{fig:alias-grounding}
\end{figure}

GRA must preserve object identity across two interfaces: the annotation LLM and perception tools exchange compact aliases, whereas training requires image-region references. We therefore convert each alias to a bounding-box token before constructing training targets. Figure~\ref{fig:alias-grounding} shows this transition for objects G and L without changing their interaction content.

\paragraph{Annotation-time} Each key object receives a short alias, such as G or L, that links the outputs of the annotation LLM and perception tools across graph-construction steps. Surround-view cameras and a LiDAR-assisted perception tool provide its 3D ego-frame state. Let $\pi_c(P_j)$ denote the two-dimensional image extent obtained by projecting the 3D box $P_j$ into camera $c$. We retain the object only if this extent intersects a valid camera image:
\begin{equation}
  \begin{aligned}
  \mathcal{O}_{\mathrm{ann}}=
  \bigl\{j:\ &\exists\, c\in\mathcal{S}_{\mathrm{ann}},\\
    &\pi_c(P_j)\cap\Omega_c\neq\emptyset\ \text{and}\ z_c(P_j)>0\bigr\},
  \end{aligned}
  \label{eq:annotation-visibility}
\end{equation}
where $\mathcal{S}_{\mathrm{ann}}$ is the surround-view camera set, $\Omega_c$ is the image extent of camera $c$, and $z_c(P_j)>0$ requires the object to be in front of that camera. Objects outside every camera view are dropped before aliases are assigned.

\paragraph{Training-time} Training may use a different camera configuration from the annotation rig, such as a single front camera or a smaller surround-view set. We therefore reproject each annotated object into the training cameras $\mathcal{S}_{\mathrm{train}}$ and keep only those still visible:
\begin{equation}
  \begin{aligned}
  \mathcal{O}_{\mathrm{train}}=
  \bigl\{j\in\mathcal{O}_{\mathrm{ann}}:\ &
    \exists\, c\in\mathcal{S}_{\mathrm{train}},\\
    &\pi_c(P_j)\cap\Omega_c\neq\emptyset\ \text{and}\ z_c(P_j)>0\bigr\}.
  \end{aligned}
  \label{eq:training-visibility}
\end{equation}
For each retained object, we resolve its alias to a normalized bounding-box token $b_j\in[0,1000)^4$ and discard references outside the training view. Annotation intermediates may contain aliases, but the resulting cognition and planning targets use bounding-box tokens. Only the reference form changes: the object-specific interaction and decision content remains intact. Section~\ref{app:gra-qa-example} shows this training-time form in complete questions, answers, and reasoning chains.

\clearpage
\endgroup
\raggedbottom
\section{Complete GRA QA Example}
\label{app:gra-qa-example}

\begin{figure}[H]
  \centering
  \includegraphics[width=\linewidth]{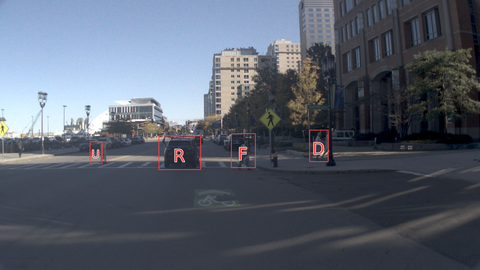}
  \caption{Front-view observation shared by all 30 QA pairs in the complete
  GRA transcript. The visual labels map to normalized boxes:
  R: (331,502),(418,629); F: (482,494),(532,622);
  D: (645,481),(684,600); U: (187,523),(213,606).}
  \label{fig:gra-qa-frame}
\end{figure}

The shared-supervision design in Sec.~\ref{sec:gra} derives cognition targets
from the same action-relevant graph used for planning. Figure~\ref{fig:gra-qa-frame}
and the transcript below show this projection for one NAVSIM frame through all
30 associated GRA QA pairs. The questions are grouped by graph node type and
follow its dependency order: scene context, grounded objects, interactions,
object-level decisions, and the ego decision. Object references in the
questions, answers, and reasoning chains use normalized bounding-box tokens of
the form
\texttt{<|box\_start|>(x1,y1),(x2,y2)<|box\_end|>}, with coordinates on a
0--1000 scale. The letter overlays in Figure~\ref{fig:gra-qa-frame} serve only
as visual lookup labels; the transcript uses the corresponding boxes directly.

\definecolor{graBlue}{HTML}{315D7C}
\definecolor{graTeal}{HTML}{24736B}
\definecolor{graPanel}{HTML}{F4F7F9}

\newcommand{\grabbox}[4]{%
  \texttt{<|box\_start|>}\allowbreak
  \texttt{(#1,#2),}\allowbreak
  \texttt{(#3,#4)}\allowbreak
  \texttt{<|box\_end|>}%
}

\newtcolorbox{graqabox}[1]{%
  enhanced,
  breakable,
  colback=graPanel,
  colframe=graBlue!55,
  colbacktitle=graBlue,
  coltitle=white,
  fonttitle=\bfseries\small,
  title={#1},
  title after break={#1\hfill{\footnotesize continued}},
  boxrule=0.45pt,
  arc=1.2mm,
  left=1.5mm,
  right=1.5mm,
  top=1mm,
  bottom=1mm,
  before skip=4pt,
  after skip=5pt,
  borderline west={1.2pt}{0pt}{graBlue}
}

\newcommand{\graqa}[2]{%
  \par\noindent
  \begin{minipage}{\linewidth}
  \raggedright
  \noindent\textcolor{graBlue}{\bfseries Q}\hspace{0.45em}#1\par
  \vspace{0.5pt}
  \noindent\textcolor{graTeal}{\bfseries A}\hspace{0.45em}#2\par
  \end{minipage}\par
}
\newcommand{\graqaseparator}{%
  \par\vspace{1pt}
  {\color{graBlue!22}\hrule height 0.35pt}
  \vspace{1pt}
}
\newcommand{\graobject}[1]{%
  \par\vspace{2pt}
  \noindent\textcolor{graBlue}{\bfseries #1}\par
  \vspace{2pt}
}

\begingroup
\linespread{0.94}\selectfont
\begingroup
\footnotesize
\raggedright

\begin{graqabox}{Scene Context}

\graqa{Please describe the current scene environment and the ego vehicle's status.}{The ego vehicle is approaching an intersection in an urban environment with multi-lane roads. The front view shows a wide intersection with buildings on the right side and open plaza area on the left. The ego vehicle is positioned at the bottom of the image, preparing to execute a right turn. Multiple vehicles are visible in the scene: vehicle \grabbox{331}{502}{418}{629} is directly ahead in the same lane, vehicle \grabbox{482}{494}{532}{622} is in the forward-left position, vehicle \grabbox{645}{481}{684}{600} is on the right side, and vehicle \grabbox{187}{523}{213}{606} is on the far left. The road structure shows a typical urban intersection with crosswalks. The ego vehicle is currently moving at low speed (3.8 m/s $\approx$ 14 km/h) with positive acceleration, indicating it is proceeding with the right turn maneuver. The future trajectory shows a smooth rightward curve consistent with a right turn navigation intent. No significant occlusion risks are apparent, but vehicle \grabbox{331}{502}{418}{629} ahead may require monitoring as the ego proceeds through the turn.}

\graqaseparator

\graqa{What traffic signs are visible in the current scene?}{A yellow diamond-shaped pedestrian crossing sign is visible in the front view on the right side of the road, indicating a crosswalk area ahead. No other clearly visible traffic signs, speed limit signs, or traffic lights are discernible in the provided images.}

\graqaseparator

\graqa{What navigation maneuver is the ego vehicle executing?}{The ego vehicle is executing a TURN RIGHT maneuver at the intersection. Based on visual observation and available information, the navigation path follows a right-turn curve. Vehicle \grabbox{331}{502}{418}{629} is positioned directly ahead on the navigation line in the same lane, approximately 15-20 meters ahead. Vehicle \grabbox{482}{494}{532}{622} is located to the left-front of the ego vehicle, and vehicle \grabbox{645}{481}{684}{600} is on the right side. The ego vehicle should maintain its current lane position and follow the right-turn trajectory while monitoring vehicle \grabbox{331}{502}{418}{629} ahead.}

\graqaseparator

\graqa{Please describe the current road congestion situation ahead of the ego vehicle.}{The ego vehicle is currently BEFORE INTERSECTION (approaching the stop line/crosswalk area). This is a queueing situation rather than congestion - vehicles are orderly positioned waiting to proceed through the intersection. In the front view and visible in the image, vehicle \grabbox{331}{502}{418}{629} is queueing directly ahead in the same lane. Vehicle \grabbox{482}{494}{532}{622} is in the left adjacent lane, and vehicle \grabbox{645}{481}{684}{600} is positioned to the right. There are approximately 3-4 vehicles visible in the immediate queue ahead of the ego vehicle (including \grabbox{331}{502}{418}{629}). The left lane also contains vehicles (\grabbox{482}{494}{532}{622} and others further ahead). The right lane appears to have vehicle \grabbox{645}{481}{684}{600}. The low speed (3.8 m/s) and positive acceleration suggest the queue is beginning to move or the ego is approaching the intersection control point. No abnormal congestion or blockage is observed - this is normal intersection queueing behavior.}

\end{graqabox}

\begin{graqabox}{Grounded Object Perception}

\graqa{What are the location, distance, speed, and heading of vehicle \grabbox{331}{502}{418}{629}?}{CAR type, coordinates [-3.26, 23.78, 0.44], distance 24.0m from ego, speed 6.17m/s, heading 0.05 rad (approximately straight). \grabbox{331}{502}{418}{629} is the leading vehicle on the navigation path, occupying the ego vehicle's driving corridor directly ahead.}

\graqaseparator

\graqa{What are the location, distance, speed, and heading of pedestrian \grabbox{482}{494}{532}{622}?}{PEDESTRIAN, coordinates [0.25, 21.85, 0.59], distance 21.9m from ego, speed 1.37m/s, heading -2.15 rad. Located near the crosswalk area forward-left of the ego vehicle, moving at walking speed.}

\graqaseparator

\graqa{What are the location, distance, speed, and heading of pedestrian \grabbox{645}{481}{684}{600}?}{PEDESTRIAN, coordinates [4.57, 23.62, 0.78], distance 24.1m from ego, speed 1.53m/s, heading -1.85 rad. Located on the right side of the turn path near the crosswalk area, moving slowly across the intersection.}

\graqaseparator

\graqa{Which obstacles in the current scene require the most attention?}{Critical obstacles are \grabbox{331}{502}{418}{629}, \grabbox{482}{494}{532}{622}, and \grabbox{645}{481}{684}{600}: \grabbox{331}{502}{418}{629} is a CAR directly ahead occupying the navigation path during the right turn, approximately 24 meters away and moving at 6.17 m/s, acting as the leading vehicle that limits the ego's maximum speed; \grabbox{482}{494}{532}{622} is a PEDESTRIAN located near the crosswalk forward-left of the ego vehicle at 21.9 meters, moving at walking speed and requiring caution as the ego traverses the crosswalk area; \grabbox{645}{481}{684}{600} is a PEDESTRIAN positioned on the right side of the turn path at 24.1 meters, also requiring monitoring to ensure safe passage through the intersection.}

\end{graqabox}

\begin{graqabox}{Object Interactions}

\graobject{Vehicle \grabbox{331}{502}{418}{629}.}

\graqa{What is the current intention of vehicle \grabbox{331}{502}{418}{629}?}{Proceeding straight through the intersection at steady speed, following the traffic flow ahead.}

\graqaseparator

\graqa{Where is vehicle \grabbox{331}{502}{418}{629} positioned relative to the ego's navigation path?}{Located in the same lane ahead of the ego vehicle, directly on the navigation path for the right turn. Based on visual analysis from the camera view, there is no lane line separation between ego and \grabbox{331}{502}{418}{629}, confirming \grabbox{331}{502}{418}{629} is the leading vehicle in the ego's lane.}

\graqaseparator

\graqa{How is vehicle \grabbox{331}{502}{418}{629} interacting with other nearby objects?}{Following vehicles ahead in the traffic queue, with vehicle \grabbox{482}{494}{532}{622} positioned to the left-front and vehicle \grabbox{645}{481}{684}{600} to the right, forming part of the multi-vehicle intersection queue.}

\graqaseparator

\graqa{What is vehicle \grabbox{331}{502}{418}{629}'s position relative to the map?}{Located in the same lane ahead of the ego vehicle, directly on the navigation path for the right turn. Based on visual analysis from the camera view, there is no lane line separation between ego and \grabbox{331}{502}{418}{629}, confirming \grabbox{331}{502}{418}{629} is the leading vehicle in the ego's lane.}

\graqaseparator

\graqa{What impact does vehicle \grabbox{331}{502}{418}{629} have on the ego vehicle?}{Acting as the leading vehicle that limits ego's maximum speed and requires maintaining safe following distance throughout the right turn maneuver.}

\graqaseparator

\graqa{What additional information is available about vehicle \grabbox{331}{502}{418}{629}?}{CAR type, moving at 6.17 m/s with heading 0.05 rad (approximately straight), distance 24.0m from ego. No brake lights visible, indicating steady movement without deceleration.}

\graqaseparator

\graobject{Pedestrian \grabbox{482}{494}{532}{622}.}

\graqa{What is the current intention of pedestrian \grabbox{482}{494}{532}{622}?}{Walking across the crosswalk or moving through the pedestrian crossing area at the intersection.}

\graqaseparator

\graqa{Where is pedestrian \grabbox{482}{494}{532}{622} positioned relative to the ego's navigation path?}{Located near the crosswalk that intersects with ego's planned right-turn trajectory through the intersection.}

\graqaseparator

\graqa{How is pedestrian \grabbox{482}{494}{532}{622} interacting with other nearby objects?}{Moving independently in the crosswalk area without direct interaction with other vehicles.}

\graqaseparator

\graqa{What is pedestrian \grabbox{482}{494}{532}{622}'s position relative to the map?}{Located in the crosswalk area visible in the image, forward-left of ego vehicle within the intersection pedestrian crossing zone, separated from ego's current lane by lane markings.}

\graqaseparator

\graqa{What impact does pedestrian \grabbox{482}{494}{532}{622} have on the ego vehicle?}{Not directly on ego's planned path but located in the crosswalk area where ego must traverse during the right turn, creating a potential conflict point that requires vigilance.}

\graqaseparator

\graqa{What additional information is available about pedestrian \grabbox{482}{494}{532}{622}?}{Pedestrian moving at 1.37m/s (walking speed) with heading -2.15 rad, distance 21.9m from ego, positioned near the crosswalk indicated by the pedestrian crossing sign.}

\graqaseparator

\graobject{Pedestrian \grabbox{645}{481}{684}{600}.}

\graqa{What is the current intention of pedestrian \grabbox{645}{481}{684}{600}?}{Walking across the crosswalk or along the intersection edge, moving slowly through the pedestrian crossing area.}

\graqaseparator

\graqa{Where is pedestrian \grabbox{645}{481}{684}{600} positioned relative to the ego's navigation path?}{Located on the right side of ego's right-turn navigation path, within the crosswalk area near the intersection.}

\graqaseparator

\graqa{How is pedestrian \grabbox{645}{481}{684}{600} interacting with other nearby objects?}{Interacting with the crosswalk environment as a pedestrian, potentially crossing the street or waiting to cross.}

\graqaseparator

\graqa{What is pedestrian \grabbox{645}{481}{684}{600}'s position relative to the map?}{Located in the crosswalk/intersection area on the right side of ego's turn path, separated from ego's lane by intersection geometry, within the crosswalk zone visible in the image.}

\graqaseparator

\graqa{What impact does pedestrian \grabbox{645}{481}{684}{600} have on the ego vehicle?}{Potential conflict during right turn as \grabbox{645}{481}{684}{600} is positioned on the right side of the turn path; requires monitoring to ensure pedestrian does not enter the turn radius or cross in front of ego.}

\graqaseparator

\graqa{What additional information is available about pedestrian \grabbox{645}{481}{684}{600}?}{Yellow pedestrian crossing sign visible in front view indicating crosswalk presence; slow walking speed of 1.53 m/s; distance 24.1 m; heading -1.85 indicating movement across the intersection area.}

\end{graqabox}

\begin{graqabox}{Object Decisions}

\graqa{What decision did the ego vehicle make regarding vehicle \grabbox{331}{502}{418}{629}, and what is the reason?}{\textcolor{graTeal}{\bfseries Think}: The ego vehicle is executing a right turn maneuver at the intersection. Vehicle \grabbox{331}{502}{418}{629} is identified as the leading vehicle positioned directly ahead on the navigation path, approximately 24 meters away, traveling at 6.17 m/s in approximately the same direction (heading 0.05). As the critical obstacle occupying the driving corridor ahead with no lane line separation, \grabbox{331}{502}{418}{629} constrains the ego vehicle's maximum allowable speed and requires the ego to maintain a safe following distance. Since \grabbox{331}{502}{418}{629} is proceeding steadily through the intersection without indications of stopping, lane changing, or deviating from the path, the appropriate decision is to follow \grabbox{331}{502}{418}{629}, adjusting the ego's speed to match the traffic flow while maintaining safe separation throughout the right turn completion.
\par\noindent \textcolor{graTeal}{\bfseries Chain}: scene\_context.navigation (TURN RIGHT) | scene\_context.summary (\grabbox{331}{502}{418}{629} directly ahead on navigation line) -> \grabbox{331}{502}{418}{629}.map\_interaction (same lane, no separation) | \grabbox{331}{502}{418}{629}.motion\_state (speed 6.17m/s, heading 0.05) -> \grabbox{331}{502}{418}{629}.ego\_interaction (leading vehicle constraint) -> Follow
\par\noindent \textcolor{graTeal}{\bfseries Answer}: Follow}

\graqaseparator

\graqa{What decision did the ego vehicle make regarding pedestrian \grabbox{482}{494}{532}{622}, and what is the reason?}{\textcolor{graTeal}{\bfseries Think}: The ego vehicle is executing a right turn at an intersection where a pedestrian crossing sign is visible, indicating a crosswalk ahead. Critical obstacle \grabbox{482}{494}{532}{622} is a pedestrian located forward-left near this crosswalk area, moving at a walking speed of 1.37m/s. Although \grabbox{482}{494}{532}{622} is not directly on the ego vehicle's planned trajectory (on\_ego\_path: false), the right turn maneuver requires the ego vehicle to traverse the crosswalk area where the pedestrian is present. Pedestrians in crosswalks have the right of way, and the combination of the turning maneuver and the pedestrian's proximity to the crosswalk creates a potential conflict situation. Therefore, the ego vehicle must exercise caution when proceeding with the right turn to monitor the pedestrian's movements and ensure safety.
\par\noindent \textcolor{graTeal}{\bfseries Chain}: scene\_context.traffic\_sign | scene\_context.navigation -> \grabbox{482}{494}{532}{622}.map\_interaction | \grabbox{482}{494}{532}{622}.motion\_state -> \grabbox{482}{494}{532}{622}.ego\_interaction -> decision
\par\noindent \textcolor{graTeal}{\bfseries Answer}: Caution}

\graqaseparator

\graqa{What decision did the ego vehicle make regarding pedestrian \grabbox{645}{481}{684}{600}, and what is the reason?}{\textcolor{graTeal}{\bfseries Think}: The front view shows a yellow diamond-shaped pedestrian crossing sign on the right side, indicating a designated crosswalk area ahead. The visual observation confirms that obstacle \grabbox{645}{481}{684}{600} is a pedestrian located in this crosswalk area on the right side of the intersection at coordinates [4.57, 23.62], approximately 24.1 meters from ego. \grabbox{645}{481}{684}{600} is moving slowly at 1.53 m/s with heading -1.85. As ego executes a right turn maneuver, \grabbox{645}{481}{684}{600} is positioned on the right side of the turn path. Although \grabbox{645}{481}{684}{600} is not directly on the planned trajectory (on\_ego\_path: false), the combination of the pedestrian crossing sign, the crosswalk location, and the slow movement of the pedestrian creates a potential interaction risk during the turn. Therefore, ego must exercise caution to monitor \grabbox{645}{481}{684}{600}'s movement closely, reduce speed if necessary, and ensure the pedestrian does not enter the turn path or cross in front of the vehicle, maintaining safety throughout the right turn maneuver.
\par\noindent \textcolor{graTeal}{\bfseries Chain}: scene\_context.traffic\_sign -> \grabbox{645}{481}{684}{600}.map\_interaction | \grabbox{645}{481}{684}{600}.motion\_state -> \grabbox{645}{481}{684}{600}.ego\_interaction -> Caution
\par\noindent \textcolor{graTeal}{\bfseries Answer}: Caution}

\end{graqabox}

\begin{graqabox}{Ego Decision}

\graqa{What is the ego vehicle's overall driving strategy considering all obstacles and navigation?}{\textcolor{graTeal}{\bfseries Think}: The ego vehicle is executing a right turn at an intersection with vehicle \grabbox{331}{502}{418}{629} as the leading obstacle directly ahead on the navigation path. The Follow decision for \grabbox{331}{502}{418}{629} requires adjusting ego speed to match traffic flow; since \grabbox{331}{502}{418}{629} is moving faster (6.17 m/s) than ego's current speed (3.8 m/s), this necessitates acceleration to maintain appropriate following distance. Pedestrians \grabbox{482}{494}{532}{622} and \grabbox{645}{481}{684}{600} are positioned near crosswalk areas on the left and right sides of the turn path respectively. While neither is directly on the ego trajectory, the Caution decisions for both indicate potential interaction risks during the right turn, requiring heightened vigilance. The trajectory intent explicitly specifies turn\_right laterally and accelerate longitudinally. Synthesizing these factors: ego must execute the right turn maneuver while accelerating to follow the leading vehicle, but remain prepared to react to the pedestrians in the crosswalk zones.
\par\noindent \textcolor{graTeal}{\bfseries Chain}: \grabbox{331}{502}{418}{629}.decision(Follow) | \grabbox{482}{494}{532}{622}.decision(Caution) | \grabbox{645}{481}{684}{600}.decision(Caution) | scene\_context.navigation(TURN\_RIGHT) | trajectory\_intent(turn\_right, accelerate) -> ego\_decision(turn\_right, accelerate)
\par\noindent \textcolor{graTeal}{\bfseries Answer}: lateral: turn\_right | longitudinal: accelerate}

\end{graqabox}

\endgroup

\let\graqa\relax
\let\graqaseparator\relax
\let\graobject\relax

\endgroup

\section{Grounded Question-Answering Evaluation Protocol}
\label{app:qa-eval}

\subsection{Evaluation Set and Judge}
\label{app:qa-set}

The class-balanced held-out set tests the cognition capabilities supporting Sec.~\ref{sec:grounding-analysis} and does not overlap with training data. Following Sec.~\ref{app:object-referencing}, all object references are converted to bounding boxes; samples without a valid box in the evaluation view are excluded before inference. The same normalized image region is rendered in each model's native box syntax.

MiniMax-M3 scores open-ended answers at temperature $0$ and is not evaluated itself. This text-only judge receives the QA type, subtype, scoring focus, question, reference, and candidate answer. Decision subtypes instead use label accuracy.

\vspace{-6pt}
\subsection{Scoring Definitions}
\label{app:qa-scoring}

Each subtype uses the scoring rule associated with the capability it measures. All reported scores range from $0$ to $10$.

\paragraph{Scene and interaction (LLM judge)} The judge scores grounding from $0$ to $2$, semantic correctness from $0$ to $4$, reasoning alignment from $0$ to $2$, and factual consistency from $0$ to $2$. Dimensions relevant to each subtype are normalized to $[0,10]$. The reference is one valid answer rather than an exhaustive checklist, so omissions and correct additions are not penalized. Contradictions, hallucinations, and vague answers lower the score.

\paragraph{Object perception (3D localization)} For perception questions with a reference 3D position, a localization score replaces the grounding dimension. It uses the Euclidean error between predicted and reference ego-frame positions,
\begingroup
\setlength{\abovedisplayskip}{6pt}
\setlength{\belowdisplayskip}{6pt}
\setlength{\abovedisplayshortskip}{3pt}
\setlength{\belowdisplayshortskip}{3pt}
\begin{equation}
  \mathrm{loc} = 2\exp\!\left(-\lVert \hat{p}-p^{*}\rVert_2 / \tau\right),
  \qquad \tau = 5\ \text{m},
  \label{eq:loc-score}
\end{equation}
\endgroup
where $\hat{p}$ and $p^{*}$ are the predicted and reference $(x,y)$ positions. Missing coordinates receive $\mathrm{loc}=0$; questions without a single target position use the judge grounding score. The localization and remaining judge dimensions are normalized to $[0,10]$. All models follow the ego-frame convention in Sec.~\ref{app:qa-prompts}.

\paragraph{Decision (label accuracy)} Object decisions use exact match over Follow, Yield, Stop, Nudge Left, Nudge Right, Overtake, and Caution. Ego decisions use the mean exact match of a lateral label (keep\_lane, nudge\_left, nudge\_right, lane\_change, turn\_left, or turn\_right) and a longitudinal label (accelerate, cruise, decelerate, decelerate\_to\_crawl, or stop). Scores are scaled to $[0,10]$, and negated mentions are ignored.

\vspace{-6pt}
\subsection{Prompts}
\label{app:qa-prompts}

\par\noindent\begin{minipage}{\linewidth}
\lstset{aboveskip=3pt,belowskip=3pt}
\paragraph{Perception coordinate prompt} Every model receives the same system prompt, which defines the ego-frame coordinates and the localization output format:

\begin{lstlisting}
You are a driving-scene 3D perception assistant. Given a single
front-camera image and the ego state, estimate the 3D position of
the queried obstacle in the EGO VEHICLE frame (units = meters):
  x = lateral offset, positive to the RIGHT
  y = longitudinal distance, positive AHEAD
  z = height above ground
  distance = sqrt(x^2 + y^2)
Answer on the first line in this exact format, then one sentence:
  coordinates [x, y, z], distance <D>m, speed <S>m/s, heading <H> rad
\end{lstlisting}
\end{minipage}\par

\par\noindent\begin{minipage}{\linewidth}
\lstset{aboveskip=3pt,belowskip=3pt}
\paragraph{Judge prompt (open-ended subtypes)} The evaluator supplies the QA type, subtype, focus, active dimensions, question, reference, and candidate answer. The common rubric below defines scoring and JSON output:

\begin{lstlisting}
Judge an autonomous-driving QA answer. Score only the active
dimensions for this subtype:
  grounding (0-2), semantic_correctness (0-4),
  reasoning_alignment (0-2), factual_consistency (0-2)
Scoring philosophy - precision over recall:
 - The reference is ONE valid answer, not a checklist. Do NOT
   penalize omissions if what the candidate says is correct.
 - Correct extra information is NOT penalized.
 - DO penalize statements that contradict the reference/scene:
   wrong decision, wrong object role, wrong signal state,
   hallucinated objects, self-contradiction. Each wrong claim
   lowers the score.
 - Vague/evasive answers that do not commit get low
   semantic_correctness, not credit for "not being wrong".
Return JSON: {"grounding":.., "semantic_correctness":..,
 "reasoning_alignment":.., "factual_consistency":..,
 "wrong_claims":[...], "rationale":".."}
\end{lstlisting}
\end{minipage}\par

\vspace{-6pt}
\subsection{Subtype-to-GRA Mapping}
\label{app:qa-subtype}

The thirteen subtypes follow the GRA graph from scene context through grounded objects and interactions to decisions.

\paragraph{Scene} Summary covers road layout, surrounding traffic, and overall context; Traffic sign covers visible signs and control states, including their absence; Navigation covers the intended maneuver, route-lane relation, and intersection context; Congestion distinguishes free flow, normal queueing, and abnormal blockage.

\paragraph{Object perception} Critical object description covers grounding, type, position, distance, speed, and heading. Critical object identification asks which objects constrain ego motion and what roles they play.

\paragraph{Object interaction} Intention evaluates a referenced object's expected behavior. Navigation interaction asks whether it affects the ego route; object interaction covers relations with other traffic participants; map interaction covers lanes, boundaries, and road structure; and ego interaction covers its effect on ego speed, path, safety margin, or right of way.

\paragraph{Decision} Object decision evaluates the response to one critical object; Ego decision evaluates the lateral and longitudinal strategy after combining object-level constraints.

\clearpage
\endgroup

\end{document}